\documentclass[11pt]{article}

\usepackage[preprint]{acl}

\usepackage{times}
\usepackage{latexsym}
\usepackage[T1]{fontenc}
\usepackage[utf8]{inputenc}
\usepackage{microtype}
\usepackage{inconsolata}
\usepackage{graphicx}

\usepackage{svg}
\usepackage{CJKutf8}
\usepackage{textcomp}  
\usepackage{scalerel}  
\usepackage[table]{xcolor}
\usepackage{tikz}
\usetikzlibrary{fit,arrows,quotes,calc,automata,positioning,backgrounds,arrows.meta,bending,shapes,snakes,matrix,shadows,fadings,decorations,decorations.text,decorations.markings,shapes.geometric,shadings,patterns,patterns.meta}
\usepackage{amsmath}
\usepackage{amsfonts}
\usepackage{amssymb}
\usepackage{comment}
\usepackage[OT2, T1]{fontenc}
\usepackage[english]{babel}
\usepackage{booktabs}
\usepackage{longtable}
\usepackage{mathtools}
\usepackage{hyperref}
\usepackage{mdframed}
\usepackage{csvsimple}
\usepackage{subcaption}
\usepackage{multirow}
\usepackage{xspace}
\usepackage{ccicons}
\usepackage{placeins}
\usepackage{epigraph}
\usepackage{tablefootnote}
\usepackage{pgfplots}
\usepackage{tcolorbox}
\tcbuselibrary{listings,breakable,skins}
\usepackage{siunitx}
\usepackage{amssymb}
\usepackage{listings}
\lstdefinestyle{mystyle}{
    keywordstyle=\color{myMAGENTA},
    numberstyle=\tiny\color{codegray},
    stringstyle=\color{myDARKBLUE},
    basicstyle=\footnotesize,
    breakatwhitespace=false,         
    breaklines=true,                 
    captionpos=b,                    
    keepspaces=true,                 
    showspaces=false,                
    showstringspaces=false,
    showtabs=false,                  
    tabsize=2
}
\tcbset{
    box/.style={
        listing only,
        listing engine=listings,
        breakable,
        sharp corners,
        colframe=black,
        coltitle=white,
        fonttitle=\bfseries\sffamily,
        boxrule=0.8pt,
        arc=0pt,
        left=1mm,
        right=1mm,
        top=1mm,
        bottom=1mm,
        listing options={
            basicstyle=\ttfamily\scriptsize,
            breaklines=true,
            breakatwhitespace=false,
            columns=fullflexible,
            keepspaces=true,
            showstringspaces=false,
            extendedchars=false,
            literate=
                {—}{{\textemdash}}1
                {‑}{{-}}1
                {–}{{\textendash}}1
                {'}{{\textquoteleft}}1
                {'}{{\textquoteright}}1
                {"}{{\textquotedblleft}}1
                {"}{{\textquotedblright}}1
                {…}{{\ldots}}1
        }
    },
    promptbox/.style={
        box,
        colback=gray!5!white, 
        colframe=myDARKBLUE,
        coltitle=white,
    }
}

\definecolor{myRED}{HTML}{D50000}
\definecolor{myBLUE}{HTML}{00FBFF}
\definecolor{myDARKBLUE}{HTML}{007ACC}
\definecolor{myGREEN}{HTML}{34B233}
\definecolor{myMAGENTA}{HTML}{C200D6}

\NewDocumentCommand{\METRIC}{m o}{%
  \tikz[baseline={($(name.base) + (0, -1.75pt)$)}]{
      \node[shape=rectangle, inner sep=2pt, inner xsep=2pt, fill=white, rounded corners=0pt, draw=black!30] (name) {
        {\tiny \bf \textsf{\hyperref[sec:metrics:metric-#1]{\textcolor{black!30}{METRIC #1}}}}};
  }\thinspace%
  \IfValueT{#2}{\ \textbf{\textsf{#2}}}%
}

\NewDocumentCommand{\LICENSE}{m o}{#1}

\usepackage{soul}
\setul{0.5ex}{0.2ex}

\newcommand{\ourmodel}{{TELL}\xspace}

\author{Aldan Creo \and Suraj Ranganath \\
    School of Computing, Information and Data Sciences\\
    University of California, San Diego\\
    United States of America \\
    \small{
        \textbf{Correspondence:} \href{mailto:research@acmc.fyi}{\texttt{research@acmc.fyi}}
    }
}

\title{Show, Don't \ourmodel:\\ Explainable AI-Generated Text Detection}

\begin{document}
\maketitle
\begin{abstract}
    Research on AI-generated text detection has presented a number of approaches to discern human from AI prose, some of which achieving high in-distribution performance. However, real-world applicability has stalled because their outputs are misaligned with the needs of users, such as professors, who are presented with a numeric score that has no attached explanation. We tackle this issue with a novel architecture, TELL, that bakes explainability from the ground-up. While our system still offers a numerical score like other detectors for comparability, TELL takes a fundamentally different approach where we aim to show the user the ``tells'' by which the model believes a text is AI or human-written, to empower the user to decide who wrote a text using their own judgment and understanding of the context of the writing and its alleged author. We train TELL on a custom SFT dataset of domain-specific authorship annotations, and further refine the system using GRPO with curriculum learning to improve performance. We achieve competitive performance with state-of-the-art detectors (AUROC \num{0.927}) while natively providing annotations that explain the basis for the detector's decision. We further evaluate the quality of our explanations using a dataset of human annotations and report a high (mean 72.3\%) win-rate on annotation concreteness, falsifiability, coherence, plausibility and grounding, allowing users to critically think and decide for themselves. Our work thus reframes the problem of AI-generated text detection in a human-centric perspective and paves the way for a new family of detectors that focus on native explainability.
\end{abstract}

\section{Introduction}
\label{sec:introduction}

\begin{figure*}[ht]
    \centering
    \begin{minipage}{0.5\textwidth}
        \centering
        \begin{subfigure}{0.5\textwidth}
            \centering
            \includegraphics[width=\linewidth]{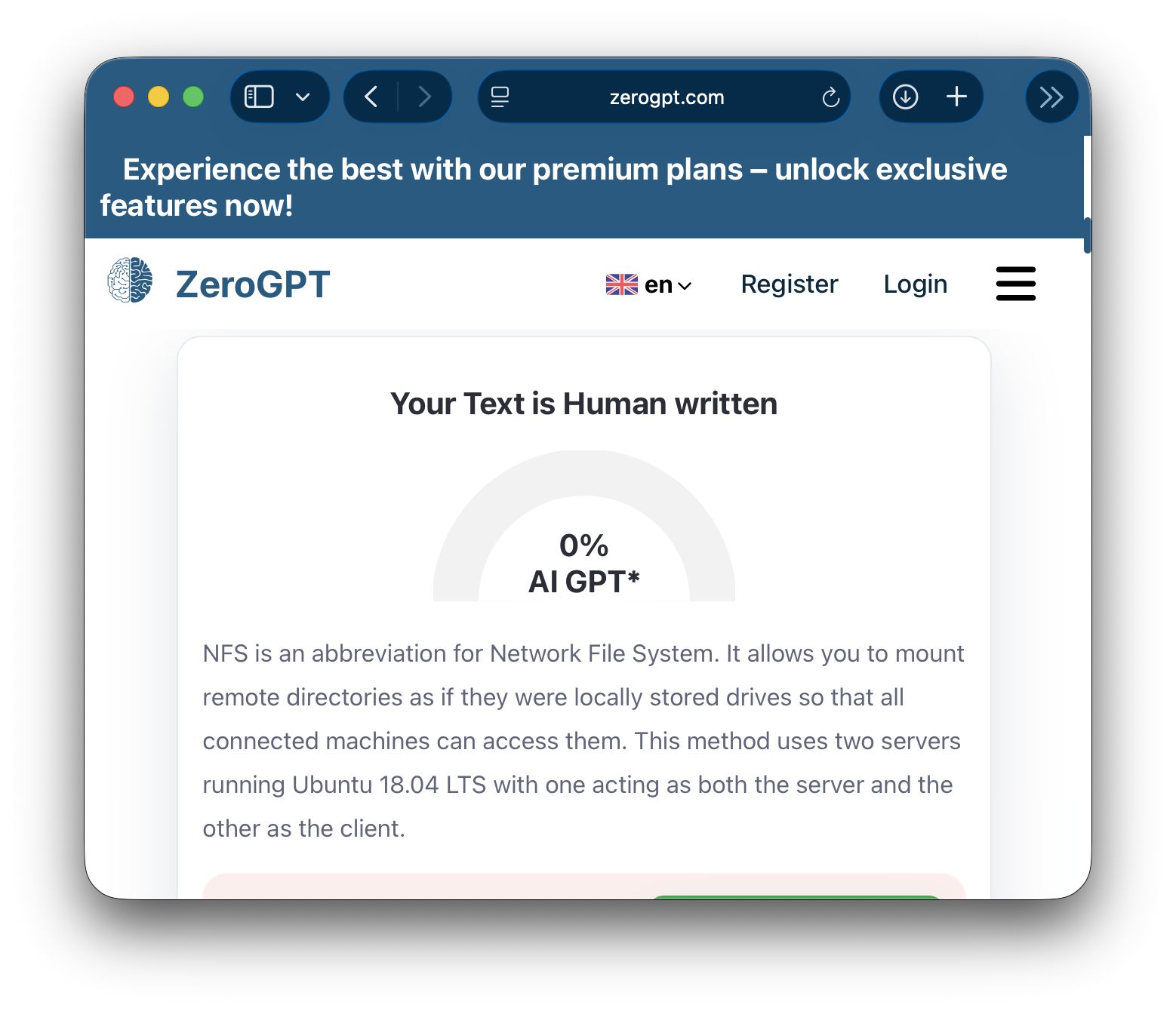}
            \caption{ZeroGPT}
            \label{fig:detectors:zerogpt}
        \end{subfigure}%
        ~
        \begin{subfigure}{0.5\textwidth}
            \centering
            \includegraphics[width=\linewidth]{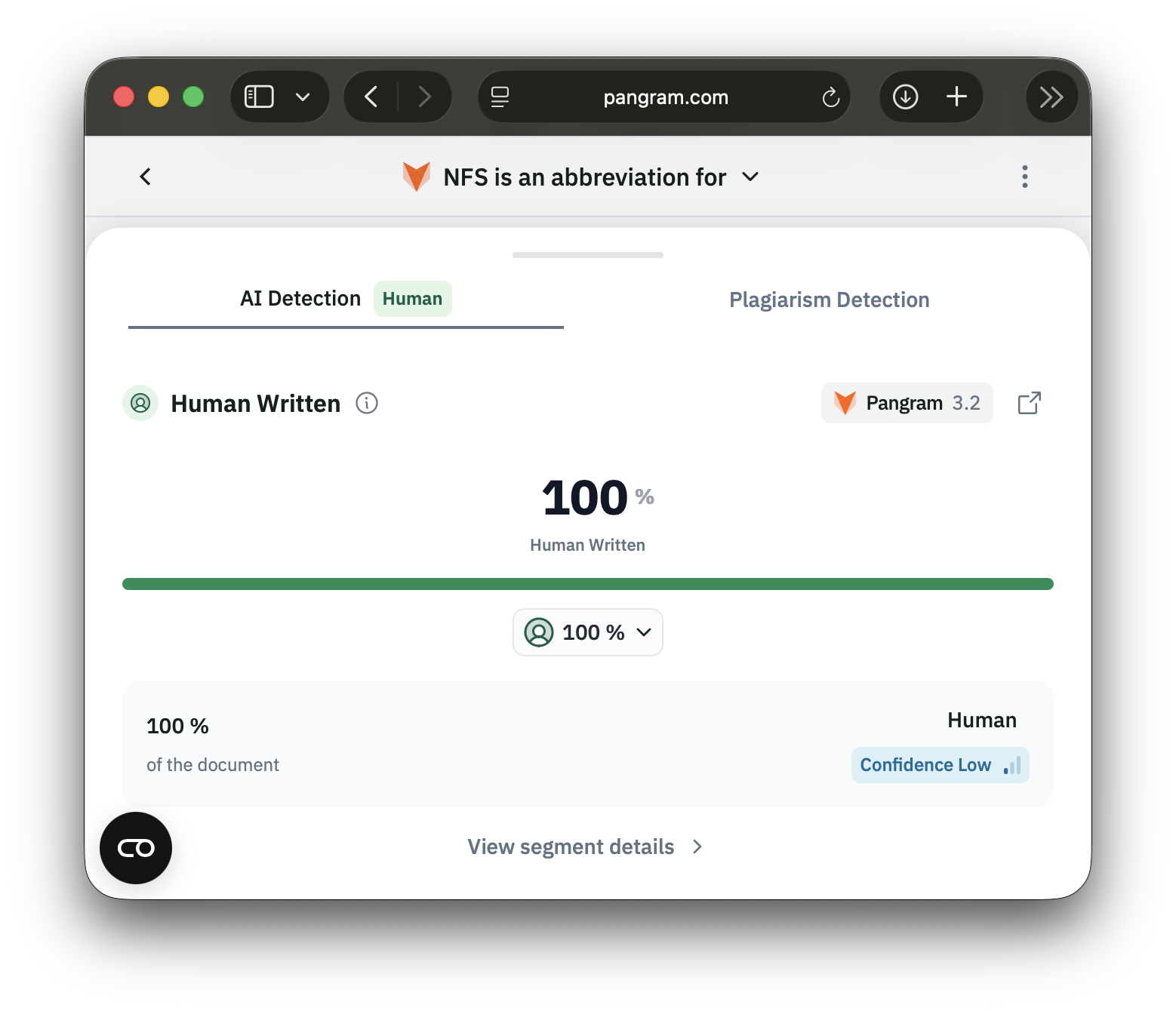}
            \caption{Pangram}
            \label{fig:detectors:pangram}
        \end{subfigure}\\
        \begin{subfigure}{0.5\textwidth}
            \centering
            \includegraphics[width=\linewidth]{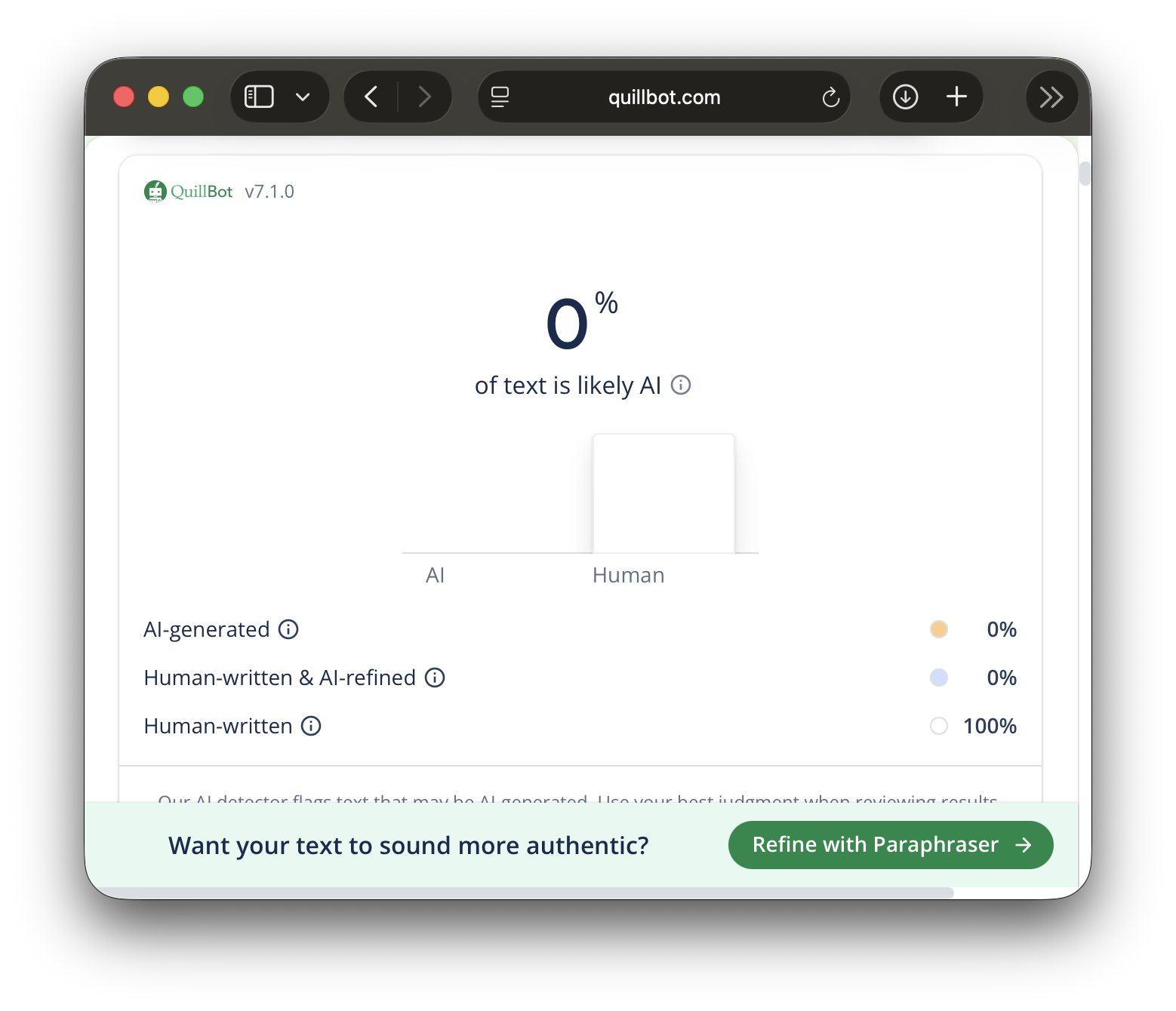}
            \caption{QuillBot}
            \label{fig:detectors:quillbot}
        \end{subfigure}%
        ~
        \begin{subfigure}{0.5\textwidth}
            \centering
            \includegraphics[width=\linewidth]{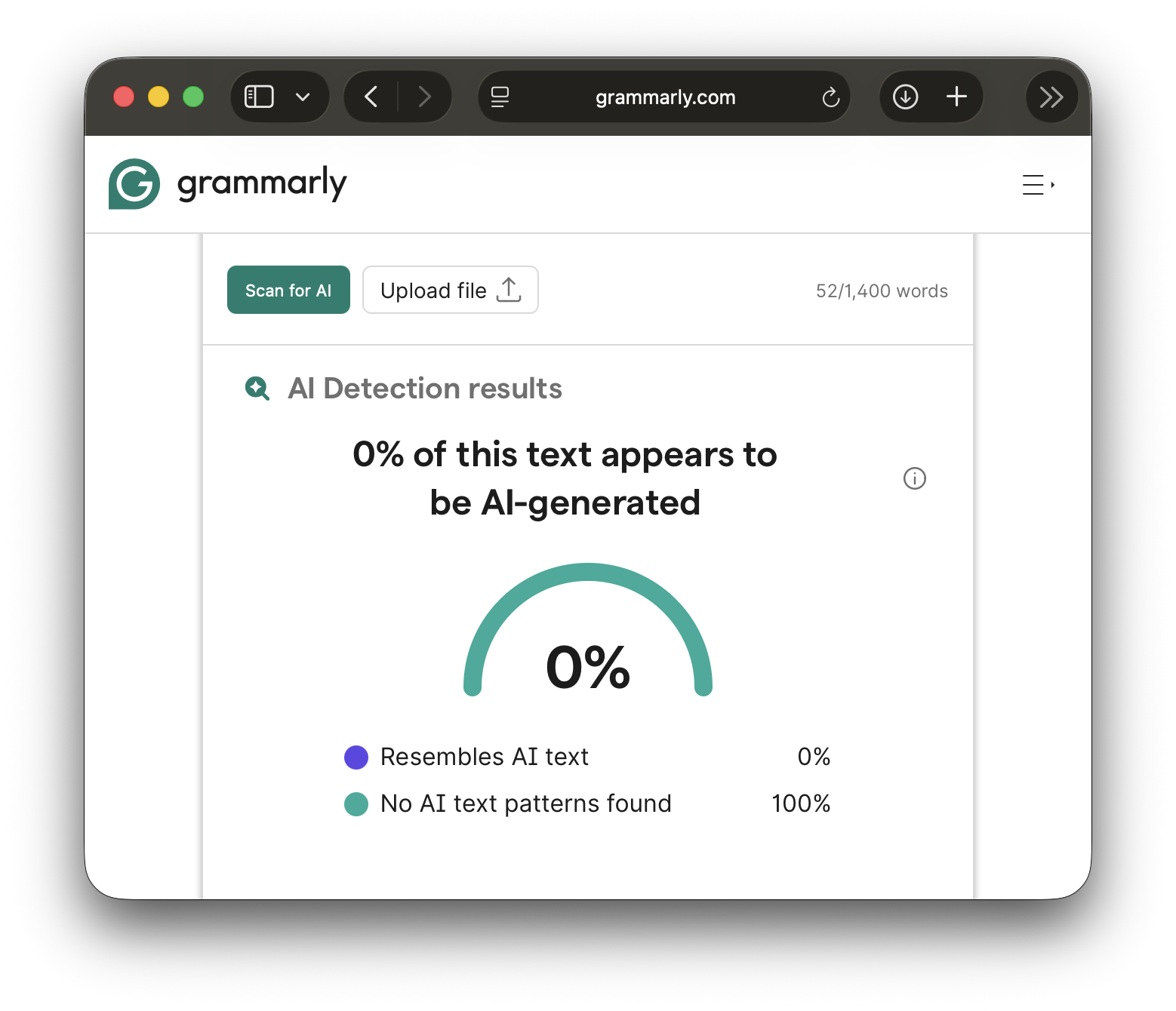}
            \caption{Grammarly}
            \label{fig:detectors:grammarly}
        \end{subfigure}
    \end{minipage}%
    \hfill
    \begin{minipage}{0.5\textwidth}
        \centering
        \begin{subfigure}{\textwidth}
            \centering
            \includegraphics[width=\linewidth]{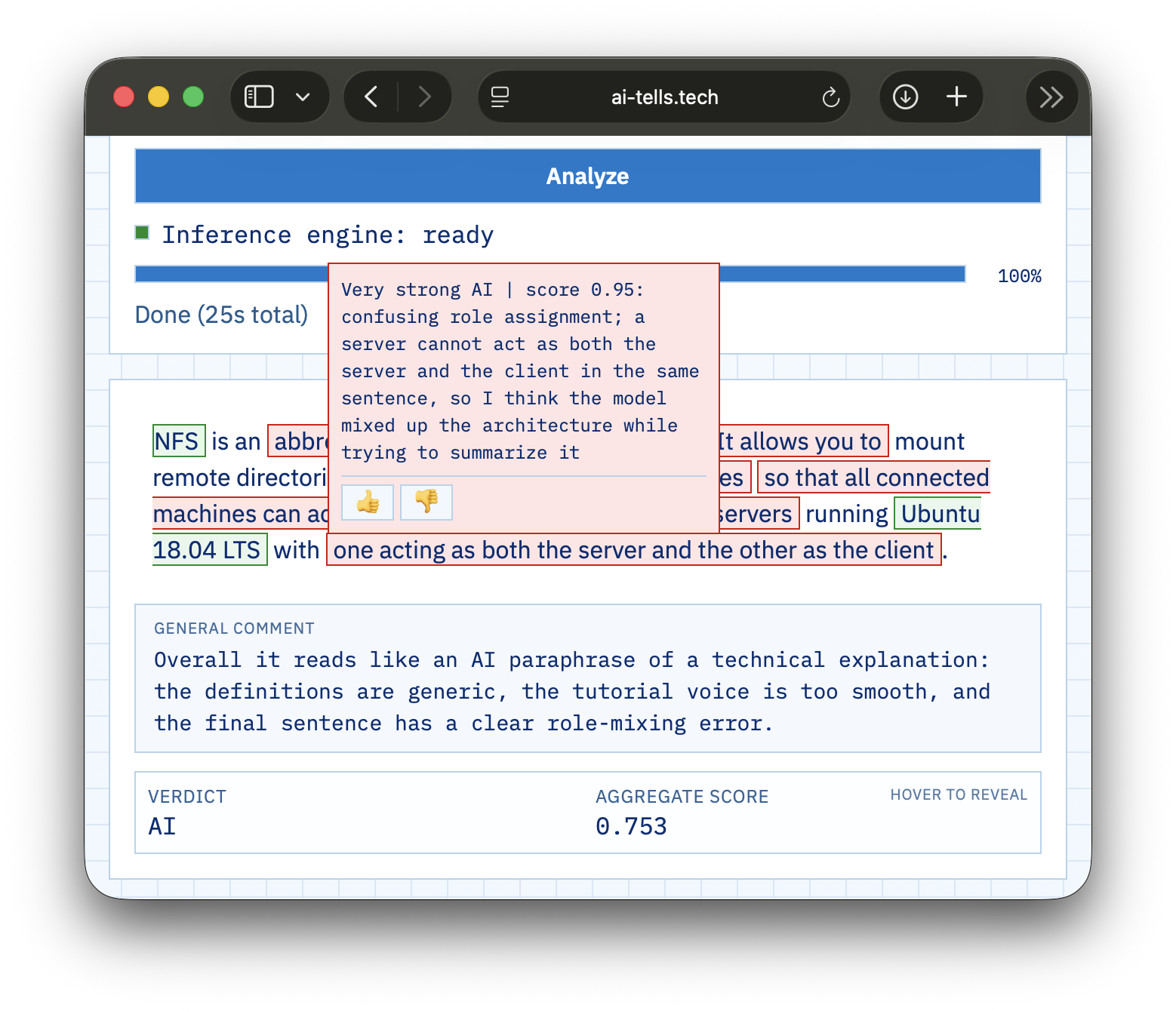}
            \caption{TELL}
            \label{fig:detectors:ours}
        \end{subfigure}
    \end{minipage}
    \caption{Existing AI-generated text detectors (e.g., \ref{fig:detectors:zerogpt}-\ref{fig:detectors:grammarly}) only give users a prediction in a binary scale. In contrast, \ourmodel~(\ref{fig:detectors:ours}) provides readable explanations for each decision, so users can understand them and make their own call.}
    \label{fig:detectors}
    \caption*{$\rightarrow$ For instance, the example above (from \citet{wangM4GTBenchEvaluationBenchmark2024}, written by \texttt{BLOOMZ}) might pass off as human-written at first glance.
    But it says that ``NFS [\dots] uses \underline{two servers} [\dots] with one acting as \underline{both} the server and the other as the client'' --- a clear contradiction (and a signal of AI generation).
    \texttt{\href{https://ai-tells.tech}{\underline{\ourmodel}}} is the only detector that can point out this kind of specific evidence and let the user judge it themselves.}
\end{figure*}


\epigraph{The purpose of computing is insight, not numbers.}{\textit{Richard Hamming}}

What does ``insight'' mean when the output of a detector is a single percentage?

Imagine a professor who submits a student's essay and receives the verdict: ``95\% AI''.
She might promptly accuse the student of cheating, having used a detector that claims to use ``comprehensive deep learning methodology, trained on extensive text collections'' \cite{ZeroGPT}. But when asked, she cannot explain her reasoning or defend it at a hearing; the detector gave her a number, but no insight.

The research community has largely focused on chasing higher accuracy. Every major detector today outputs a score as a verdict (Figure~\ref{fig:detectors}), and the literature is saturated with models claiming to surpass existing ones. So if accuracy is the answer, why does public trust in detectors remain fragile?

\subsection{A ``crisis of trust'' in detectors}

When reading the literature on AI-generated text detection, it is possible to find a vast number of papers reporting near-perfect results on their test sets, with a field saturated with new models and techniques claiming to surpass existing ones \cite[inter alia]{adamGPTZeroRobustDetection2026,emiTechnicalReportPangram2024,hansSpottingLLMsBinoculars2024,mitchellDetectGPTZeroShotMachineGenerated2023,koikeOUTFOXLLMgeneratedEssay2023,huRADARRobustAIText2023,liMAGEMachinegeneratedText2023,zhanG3DetectorGeneralGPTGenerated2023,liuArguGPTEvaluatingUnderstanding2023,houSemStampSemanticWatermark2023,solaimanReleaseStrategiesSocial2019}.

One might naturally assume that detecting AI-generated text is ``a solved problem'', and democratizing detectors is enough to respond to the concerns of, for instance, 76\% of Americans who say it is ``extremely or very important to be able to tell if pictures, videos, and text were made by AI or people'' \cite{kennedyHowAmericansView2025}, 90.4\% of people in the UK who are concerned about deepfakes \cite{sippyDeepfake8Create2024}, or the 87\% of EU27 citizens ``moderately or highly worried'' about fake content created by AI \cite{europeanparliamentParlemeterEPAutumn2026}.

However, performance reported by the original articles does not always stand the test of time and community scrutiny. For example, while the original Fast-DetectGPT \cite{baoFastDetectGPTEfficientZeroShot2024} paper reports a \num{0.9887} AUROC, \citet{tuftsPracticalExaminationAIGenerated2024} report a lower \num{0.8405}, \citet{wuDetectRLBenchmarkingLLMGenerated2024} \num{0.5533}, and \citet{chenDivScoreZeroShotDetection2025} \num{0.4632}.
This might be due to a multitude of reasons, such as selective reporting, dataset curation or other methodological choices that inflate results, datasets turning increasingly challenging over time, generator models progressively reducing the human-AI language modeling gap, or other ``perverse incentives'' that can shape the publication dynamics of a ``hot'' research community \cite{ioannidisWhyMostPublished2005}.

In fact, recent public incidents show how AI-text detectors fail in real settings, including false accusations and inconsistent results across tools \citep[inter alia]{chenWritersAreGoing2026,mingayBeingFalselyBranded2026,universityofsandiegolegalresearchcenterGenerativeAIDetection2025,nprTeachersAreUsing2025,mathewsonCaliforniaCollegesPay2025,weinbergAIChatGPTTurnitin2025,arditoContraGenerativeAI2023,scarfeRealworldTestArtificial2024,fowlerWeTestedNew2023}, as detectors tend to struggle with out-of-distribution examples such as unseen domains, obfuscated or paraphrased texts, or texts produced by newer models \cite[inter alia]{pudasainiWhyAIGeneratedText2026,shekharDoPEDecoyOriented2026,ranganathStealthRLReinforcementLearning2026,davidAuthorMistEvadingAI2025,ayoobiESPERANTOEvaluatingSynthesized2024,burgerCanAIRecognize2025,huangAreAIGeneratedText2024,duganRAIDSharedBenchmark2024,creoSilverSpeakEvadingAIGenerated2024,weber-wulffTestingDetectionTools2023,sadasivanCanAIGeneratedText2023}.
In fact, a part of the student body has grown increasingly worried of being mistakenly accused of academic integrity violations --- with some talking about a ``witch hunt'' in the educational system --- and thus try to, among other things, add typos to their work, ``dumb down'' their writing, or use online ``AI humanizers'' that claim to make AI-generated text appear more human-like (even if their text was human to begin with) \cite{watersTypoVibeShift2026,kingkadeCollegeStudentsAre2026,friedmanAIDetectionAuthors2026,scibiliaGuiltyProvenHuman2025}.

We are not claiming that all detectors are inaccurate, but rather that even the best ones can fail in practice. For instance, back in 2023, several news outlets reported that detectors were incorrectly flagging the US Constitution as AI-generated \citep[inter alia]{universityofsandiegolegalresearchcenterGenerativeAIDetection2025,woodAIDetectorsAre2024,naprysYourEssayWas2024,jiangAIgeneratedContentActually2023,edwardsWhyAIWriting2023}. This was fixed by subsequent work, but the damage was already done; currently, a substantial part of the public is skeptical about the reliability of detectors or simply avoids using them altogether \citep{shepherdGenerativeAIMisuse2025,roeUnderstandingStudentAcademic2024,gengDetectabilityLLMGeneratedText2025}.

We call this a ``crisis of trust'' in AI-generated text detection: one visible failure can damage trust far more than many correct predictions can rebuild it.
And the mistrust is aggravated when the only output of the model is a numeric score and users can't understand \textit{why} a model misclassified a text --- in the case of the US Constitution example, users could arguably have been more forgiving if the detector had explained that ``I know this text'' and thus an AI could have memorized it. Instead of ``this detector doesn't work'', users might think, ``this is an edge case, so it's normal that the detector got it wrong''.
And yet, even though public discourse and the scientific literature have repeatedly pointed out that explainability is a ``major issue'' \cite{thomsonHowAIDetection2025,anlenTRIEDTrulyInnovative2025,sahaAlmostAIAlmost2025,jiDetectingMachineGeneratedTexts2024}, to the best of our knowledge, existing work has failed to frame AI-generated text detection around explainability as its core design goal.


\subsection{Building trust with explainability}

\begin{figure*}[t]
    \centering
    \input{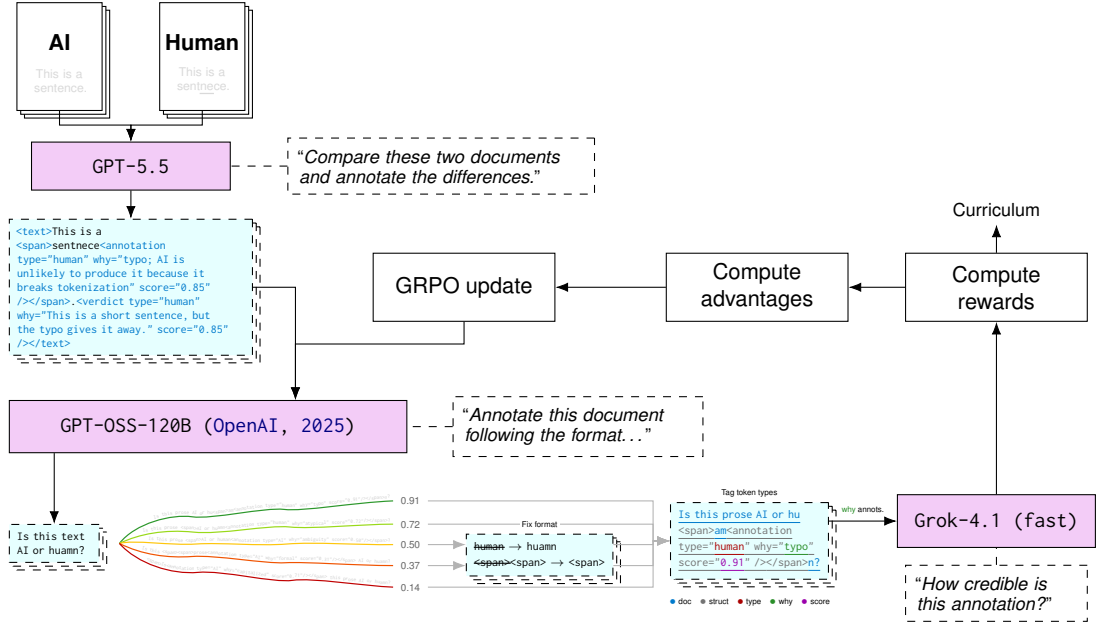}
    \caption{\textbf{Our approach.} We start by generating a dataset of annotation examples that we use to fine-tune a pretrained LM. That intermediate model is not able to reliably identify AI-generated text (AUROC \num{0.638}, TPR@1\%FPR \num{0.0}), but has learned the general task setup. We further refine it with reinforcement learning (GRPO) till convergence. The resulting model is reasonably accurate (AUROC \num{0.927}, TPR@1\%FPR \num{0.638}) --- but most importantly, explainable by design.}
    \label{fig:model_architecture}
\end{figure*}



With \ourmodel, our goal is not to claim near-perfect detection, since (1) all detectors can fail and need human oversight; (2) many need retraining as new models are released; (3) we find that related work claims high performance that does not replicate in our experiments or (arguably) the real world; and most importantly, (4) focusing on accuracy alone misses the point of why people use AI detectors, i.e., to make decisions in the real world. Instead, we design an architecture that returns a verdict together with specific evidence, so users can see why the model made that decision and check it themselves.

In fact, we believe that the utility of \ourmodel~goes beyond just being a more explainable detector. While several authors have reported that humans are not accurate AI detectors \citep{fiedlerHumansIdentifyAIgenerated2025,cookeGoodCoinToss2025,chengAbilityAIDetection2025,frankRepresentativeStudyHuman2023}, this assumption has been challenged \citep{russellPeopleWhoFrequently2025,milickaHumansCanLearn2025}. Recent work has shown that humans that undergo some kind of (possibly informal) training process can show high degrees of accuracy. In this context, \ourmodel~can help users learn, since it gives detailed explanations that allow them to generate their own mental model of what AI-generated text looks like, and thus become better detectors themselves.

Therefore, our contributions are as follows:
\begin{itemize}
    \item We propose a novel architecture, \ourmodel, that puts explainability at the core of AI-generated text detection.
    \item \ourmodel~is trained on a custom SFT dataset of domain-specific authorship annotations, and further refined using GRPO with curriculum learning to improve performance.
    \item \ourmodel~achieves state-of-the-art performance (Section~\ref{sec:results:detector_benchmark}) --- but most importantly, \ourmodel~gives users high-quality, explained evidence (Section~\ref{sec:results:annotation_quality}) that allows them to understand the reasons behind the verdict, and gives them the ability to judge on their own.
    \item \ourmodel~may serve as a didactic tool to train users in AI-generated text detection, as prior research has shown that after training, humans may be accurate detectors themselves.
    \item We openly release our code, data and weights (\url{https://github.com/ACMCMC/TELL}) to facilitate research in explainable AI-generated text detection, as well as an interface (\url{https://ai-tells.tech}) to make it accessible.
\end{itemize}

\begin{table*}[t]
\centering
\footnotesize
\begin{tabular}{@{}p{0.32\linewidth}p{0.11\linewidth}p{0.36\linewidth}r@{}}
\toprule
Source & License & Primary role in corpus & Rows \\
\midrule
RAID~\citep{duganRAIDSharedBenchmark2024} & \LICENSE{MIT} & Multi-domain and adversarial benchmark & \num{7654920} \\
COLING 2025~\citep{alamProceedings1stWorkshopGenAI2025} & \LICENSE{Apache-2.0} & Multi-domain (aggregates \citet{liMAGEMachinegeneratedText2023,guoHowCloseChatGPT2023,wangM4GTBenchEvaluationBenchmark2024}) & \num{872525} \\
OpenLLMText~\citep{chenOpenLLMTextDataset2023} & \LICENSE{\ccby~4.0} & Web text with human/AI labels & \num{344530} \\
AuTexTification~\citep{sarvazyanOverviewAuTexTificationIberLEF2023} & \LICENSE{\ccbyncsa~4.0} & Multi-domain detection data across social, review, news, legal, and how-to domains & \num{107868} \\
OUTFOX~\citep{koikeOUTFOXLLMgeneratedEssay2023} & \LICENSE{Apache-2.0} & Student essays with LLM generations and adversarially generated attacks & \num{63600} \\
Pangram EditLens~\citep{thaiEditLensQuantifyingExtent2025} (\texttt{human\_written} and \texttt{ai\_generated} rows) & \LICENSE{\ccbyncsa~4.0} & Human-written and AI-generated corpus & \num{51115} \\
DAIGT-v2~\citep{kleczekDAIGTV2Train2023} & \LICENSE{MIT} & Student-essay style human/AI texts & \num{44864} \\
AI-and-Human-Generated Text~\citep{theocharopoulosWhoWritesReview2024} & \LICENSE{MIT} & Academic abstracts & \num{28662} \\
Ghostbuster Essay~\citep{vermaGhostbusterDetectingText2023} & \LICENSE{\ccby~3.0} & Human and AI-written essays & \num{7000} \\
ArguGPT~\citep{liuArguGPTEvaluatingUnderstanding2023} & \LICENSE{\ccby~4.0} & GPT-generated argumentative essays & \num{4038} \\
\midrule
\textbf{Total} &  & \textbf{Our RL corpus} & \textbf{\num{9179122}} \\
\bottomrule
\end{tabular}
\caption{Source datasets in our training corpus after filtering and normalization.}
\label{tab:dataset_sources}
\end{table*}

\subsection{Related work}
\label{sec:related_work}

To the best of our knowledge, no existing work explicitly frames AI-generated text detection as an explainability problem, but there are some works related to this view.

\paragraph{Reasoning in LLMs.} The shift from direct generation to reasoning-based responses in LLMs mirrors what we propose. Reasoning models were motivated (among other factors) by the desire to make them more interpretable by showing how they arrive at their conclusions \citep{openaiOpenAIO1System2026,guoDeepSeekR1IncentivizesReasoning2025,weiChainofThoughtPromptingElicits2023}. We take inspiration from that shift, moving from a direct classification to one based on interpretable evidence.

\paragraph{Prompt inversion.} \citet{chenIPADInversePrompt2025} proposed a method to detect AI-generated text by inverting the prompt and checking if the model can reconstruct the original input. However, we argue that (1) this method is not designed to provide explainability to users, and (2) it may mislead them by ``finding a generating prompt'' even when the text is human-generated --- which ``primes'' them into believing whatever the model outputted.

\paragraph{Attribution.} Some works have explored attribution methods to find which parts of a text are most indicative of AI generation. For example, prior work used SHAP \cite{najjarLeveragingExplainableAI2025}, while \citet{yanExplainableFrameworkAssisting2025} used Layer Integrated Gradients to measure neuron contributions via gradient attribution. But all these methods are post-hoc: the model produces a score first, and then attribution tries to explain it after the fact, which is non-trivial and can easily lead to unfaithful explanations. In contrast, we build explainability from the ground up; it's inherent to our model. NOTAI.AI \citep{breneurNOTAIAIExplainableDetection2026} provides a similar approach by integrating curvature-based signals with neural and stylometric features for explainability, though this is done through XGBoost and SHAP attribution given to an LLM to write the explanations \textit{a posteriori}. Similarly, \citet{yuanEMMMExplainMe2025} proposed EMMM, a framework where they use Faith-SHAP to select salient tokens and convert them into natural language using templates. However, their work focuses on customer service chatbot interactions, and the explanations are also post-hoc.

\section{Methods}
\label{sec:methods}

In this section, we describe how we collect data and train and evaluate \ourmodel.

\subsection{Datasets}
\label{sec:datasets}

We use two sources of data at different stages:

\paragraph{SFT data.}
We first train our model on the task mechanics with SFT on annotated span examples. To the best of our knowledge, no dataset exists that provides AI/human annotated text with span-level annotations and natural language explanations. Therefore, we built our own dataset on top of the EditLens~\citep{thaiEditLensQuantifyingExtent2025} dataset, which pairs human text with AI-edited variants. We use \texttt{GPT-5.5} to compare each pair and generate span-level annotations, for both the human and the AI-edited document. We limit this to \num{2000} examples, partly because generation is expensive, but mostly because the goal of this stage is just to teach the model the tell-annotation format, not to build an accurate detector yet.
Additionally, we combine it with the dataset by \citet{russellPeopleWhoFrequently2025}. It contains 300 documents of human annotators indicating not only whether they think a text is human or AI-generated, but also their natural text commentary. We take 100 elements from it to generate additional SFT examples with \texttt{GPT-5.5} and \texttt{GPT-5.4}, using a prompt where we ask the model to annotate the document based on the final commentary (since the dataset doesn't contain annotated spans, which is a key motivation for our work). Each example contains the verdict of 5 annotators, but we filter the commentaries that are shorter than 50 words to ensure that the model has enough information to generate annotations from. This results in \num{316} additional SFT examples. We include the prompts in Appendix~\ref{sec:appendix:prompts}.

\paragraph{Reinforcement learning (RL) and test data.}
We build a unified dataset aggregating 10 public sources spanning 15 domains (e.g., academic abstracts, creative writing, news, student essays\dots) for a total of 9.2M rows (Table~\ref{tab:dataset_sources}).
However, our source datasets have very different sizes, so if we were to sample uniformly, our model would learn to exploit the features in e.g. RAID rather than generalizing.
So instead of drawing proportionally or equally (ignoring dataset size), we (1) define strata based on the combination of dataset and domain, and (2) allocate examples per stratum by $\sqrt{\text{stratum size}}$ as the geometric mean of the two.
We generate three splits with the same policy on scale, and ensure a 50/50 balance of AI/human examples.

The validation and test sets contain 5,000 examples. We chose this number based on statistical power: when collecting data, we assumed approximately 10 detectors in our comparison, for a total of $\binom{10}{2} = 45$ pairwise comparisons.
Using the DeLong variance formula and Benjamini--Hochberg FDR correction at $q = 0.05$, this gives 86--99\% power to detect a $\Delta = 0.04$ AUROC gap, the smallest difference we consider practically meaningful for a deployed detector.
We do this to achieve a balance between minimizing computational costs and ensuring that we have sufficient statistical power to assess model performance.

\begin{table}[]
\centering
\footnotesize
\setlength{\tabcolsep}{4.2pt}
\begin{tabular}{@{}l l r@{}}
\toprule
Detector & AUROC (95\% CI) & TPR@1\%FPR \\
\midrule
\ourmodel~(ours) & \textbf{\num{0.927} [\num{0.919}, \num{0.935}]} & \textbf{\num{63.8}} \\
MAGE & \num{0.913} [\num{0.904}, \num{0.922}] & \num{4.2} \\
Pangram-EditLens & \num{0.911} [\num{0.903}, \num{0.919}] & \num{58.3} \\
Fast-DetectGPT & \num{0.861} [\num{0.850}, \num{0.872}] & \num{59.0} \\
ArguGPT & \num{0.828} [\num{0.816}, \num{0.840}] & \num{43.3} \\
T5Sentinel & \num{0.802} [\num{0.790}, \num{0.814}] & \num{17.5} \\
DetectLLM-NPR & \num{0.782} [\num{0.769}, \num{0.795}] & \num{32.0} \\
OpenAI RoBERTa & \num{0.777} [\num{0.764}, \num{0.789}] & \num{33.1} \\
AIGC MPU & \num{0.774} [\num{0.761}, \num{0.787}] & \num{11.6} \\
DetectLLM-LRR & \num{0.763} [\num{0.749}, \num{0.776}] & \num{27.2} \\
LogRank & \num{0.757} [\num{0.744}, \num{0.771}] & \num{23.2} \\
RADAR & \num{0.744} [\num{0.730}, \num{0.758}] & \num{1.3} \\
ChatGPT-D & \num{0.697} [\num{0.682}, \num{0.711}] & \num{16.6} \\
Binoculars & \num{0.616} [\num{0.601}, \num{0.632}] & \num{1.4} \\
DNA-GPT & \num{0.581} [\num{0.566}, \num{0.595}] & \num{0.0} \\
PHD RoBERTa & \num{0.521} [\num{0.505}, \num{0.537}] & \num{4.6} \\
\bottomrule
\end{tabular}
\caption{Comparison of detection methods. \ourmodel~achieves the best scores (metrics: higher is better).}
\label{tab:main_detector_benchmark}
\end{table}

\subsection{Model architecture and training.}
\label{sec:model_architecture}

\subsubsection{Supervised fine-tuning}
\label{sec:sft}

We used the examples generated (Section~\ref{sec:datasets}) for the supervised fine-tuning (SFT) step on \texttt{GPT-OSS-120B} (see Appendix~\ref{sec:appendix:sft_details} for details). We also include the real human comments by \citet{russellPeopleWhoFrequently2025}.

Additionally, we include an extra CE loss on ``hint'' following: during the RL stage, we artificially inject the document label in some of the rollouts, so that the model has signal on what the correct annotation should be. Otherwise, the variance of the reward might be too small and the model might collapse, e.g. if the model is strongly confident on a wrong label. For that, we train on dual pairs of examples of text where there is a hint (``Text origin is AI/human'') in the reasoning section, and the only token that is optimized is the final verdict token (either ``AI'' or ``human''). This way, the model learns to associate the hint with the correct label, and we leverage this during the RL stage to provide a stronger learning signal.

\subsubsection{RL training}
\label{sec:rl}

After SFT, we further refine the model with GRPO \citep{shaoDeepSeekMathPushingLimits2024}, where we aim to progress from a model that is familiar with the task format, though not accurate, to one that can reliably identify AI-generated text and provide high-quality annotations.
We describe our main methodological choices (which substantially deviate from a standard GRPO implementation) in the following paragraphs.

\paragraph{Curriculum.}
\label{sec:curriculum}
Not all training documents are equally informative at every stage of training.
Documents where the model always fails or succeeds have no group variance (thus give zero GRPO gradient).
We therefore implement a curriculum \citep{bengioCurriculumLearning2009} that dynamically prioritizes the strata where documents show the maximum reward variance, in line with the approach by \citet{emiTechnicalReportPangram2024}, who showed it can be highly effective to maximize model learning in developing the Pangram detector.

Our training data is heterogeneous, spanning multiple datasets and domains (see Section~\ref{sec:datasets}), so we first partition documents into strata by dataset and domain.
For each stratum, we maintain an EMA of the classification reward as a proxy for difficulty.
We then sample strata using a Gaussian curriculum window centered at a target difficulty $\tau$, which linearly ramps from $\tau_\text{start} = 0.35$ to $\tau_\text{end} = 0.70$ over the first 50 training steps; the model starts training on moderately hard strata and as training progresses, it is given harder examples.
Within each selected stratum, we use a UCB exploration term to also ensure that we have coverage of less-visited strata (to avoid local optima).

\paragraph{Replay.}
\label{sec:replay}
We additionally maintain a cache of successful rollouts (up to \num{6000} entries), from which we sample a growing fraction of each batch (starting at 35\%, ramping to 50\% by step 80).
This way, we stabilize the gradient signal by mixing fresh rollouts with those where reward variance was maximized (i.e., the model had the most learning potential), and we also increase the effective batch size without additional expensive decoding.




\paragraph{Format collapse.}
\label{sec:format_collapse}
We found that the model can sometimes collapse to a degenerate behavior, where it, for example, ``corrects'' grammatical errors in the input by generating the corrected version instead, or simply repeats structural tokens, or hallucinates a false output. To address this, we implemented a format-fixing pipeline that detects and corrects wrong-format rollout (up to a 10\% difference), and we use the format-fixed version for GRPO updates, and we then apply an additional cross-entropy loss gradient up to the corrected version's doc-copy and structural tokens (the annotations are zeroed out), so that the model is reinforced to produce correctly-formatted text. Initial experiments where we applied the reward directly with a reward of 0 for wrong-format rollouts led to a complete collapse, since differences are frequently small (e.g., a missing comma), and GRPO was negatively reinforcing all tokens in those wrong rollouts when the culprit was only a small fraction of them.


\paragraph{Per-Token Advantage Decomposition.}
\label{sec:ptad}
In our output, tokens perform different roles depending on their position. For instance, in
\begin{quote}
\texttt{<text>This <span>is<annotation type="human" why="reason" score="0.12" /></span> an\dots}
\end{quote}
the first token \texttt{<text>} is structural --- it always appears at the beginning. Then, \ourmodel~should start copying the original document text --- either writing the same token as in the original document (\texttt{This}), or choosing to open a span (\texttt{<span>}). After opening a span and copying the tokens inside, \texttt{<annotation type="} is a structural token again, but then the model needs to write the annotation type (\texttt{human} or \texttt{AI}), another structural token, and a potentially long explanation (\texttt{reason}), etc.

If we were to apply a single scalar advantage across all tokens, the model would receive the same learning signal for all of them, which in practice leads to format collapse. In all rollouts, better and worse, the model should still reproduce the original document text --- and if we were to negatively reward in some rollouts, then it would get a contradictory signal (``you should only write the original document when you're correct''). The same reasoning applies to structural tokens and annotation tokens. The ``tasks'' performed by different tokens are different, and thus they require different learning signals.

Therefore, we assign each token an advantage based on its structural role, using independent reward pools per token type.
We always give document-copy tokens zero advantage: the model should reproduce the original text regardless of output quality, and rewarding or penalizing this would introduce contradictory signal\footnote{We also experimented with a small fixed advantage, but found it to be of no benefit as we already apply format fixing.}.
We also give structural tokens a small fixed positive advantage to reinforce the adherence to our format.



\paragraph{Reward functions.}
\label{sec:reward_functions}
We only optimize rollouts with a valid format, and we assign independent rewards to each token type (Section~\ref{sec:ptad}).
For type tokens, we model the reward as the product of the rubric credibility and the label alignment: $c \cdot (+1)$ if the annotation type matches the document label, and $c \cdot (-1) + 1$ if it does not. This way, a high-credibility but of the label opposite to the document (i.e., the model identified high quality evidence that happens to contradict the overall label) receives only a small penalty, while a low-credibility wrong-type annotation receives the strongest negative signal.
For the annotation and verdict explanation tokens, the reward is the product of the rubric credibility and a quality gate based on length and repetition across annotations. We get the credibility score from a frozen LLM judge, \texttt{Grok-4.1-Fast} (see prompt in Appendix~\ref{sec:appendix:prompts}).
For the score tokens, we reward as $1 - |\hat{s} - c|$, where $\hat{s}$ is the model's written score and $c$ is the rubric credibility; this teaches the model to calibrate its confidence to match the judge's.
We use these rewards on GRPO normalization with separate pools for each token type, e.g., a rollout can receive a strong positive signal on its \texttt{type} token and also receive a negative signal on its \texttt{why} tokens, so that the model can learn classification and explanation quality at the same time.


\section{Results and discussion}
\label{sec:results}

\begin{table}[t]
\centering
\begin{tabular}{l r}
\toprule
{Judge} & {Win rate (\%) [95\% CI]} \\
\midrule
\bfseries Panel mean & \bfseries \num{72.3} [\num{68.3}, \num{76.2}] \\
\midrule
  GPT-5.4-mini & \num{78.3} [\num{73.9}, \num{82.4}] \\
  Gemma 4 26B & \num{67.5} [\num{62.6}, \num{72.1}] \\
  DeepSeek V4 Flash & \num{75.3} [\num{70.8}, \num{79.5}] \\
  Nemotron Super & \num{66.3} [\num{61.5}, \num{70.8}] \\
  GPT-OSS 120B & \num{74.1} [\num{69.5}, \num{78.4}] \\
\bottomrule
\end{tabular}
\caption{Win rate vs.\ human experts. 95\% CIs: document-level bootstrap ($B=\num{10000}$).}
\label{tab:winrate_compact}
\end{table}

\begin{figure}[t]
    \centering
    \includegraphics[width=1.0\linewidth]{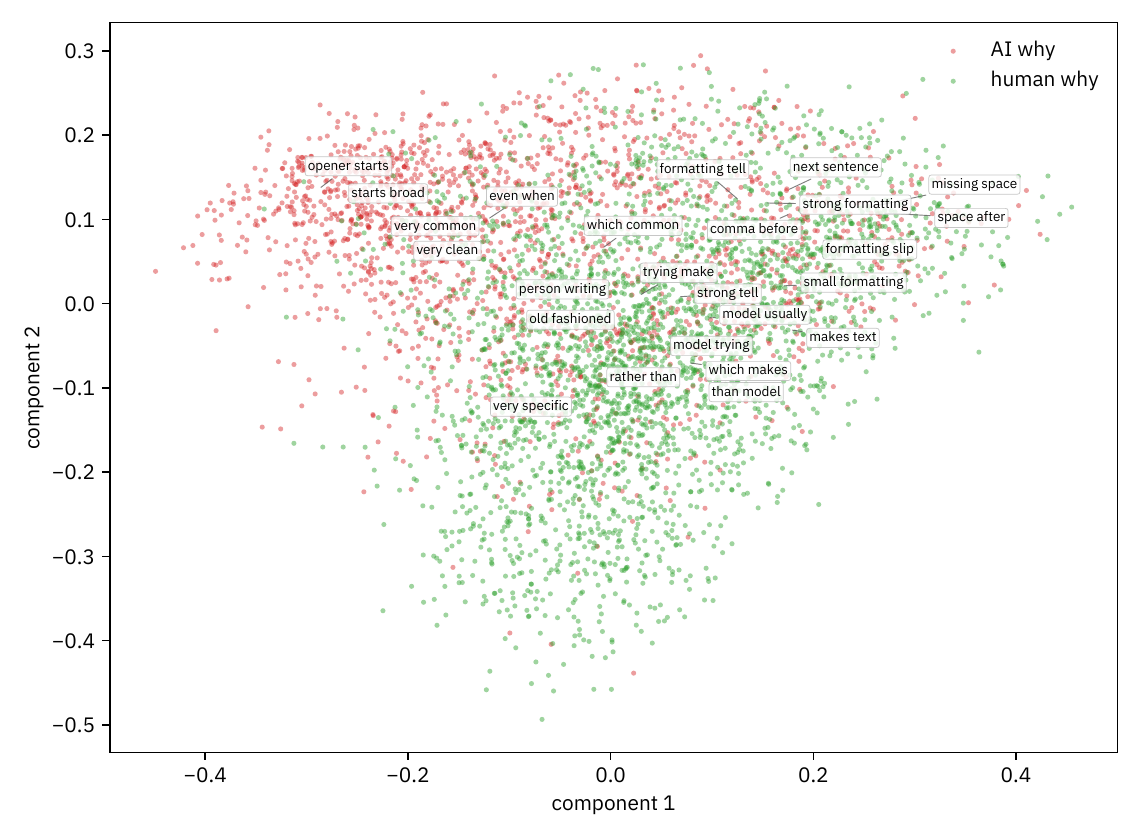}
    \caption{\textbf{What \ourmodel~writes.} There is human/AI separation, with specific ``attributes'' relating to each (e.g., ``very common'', ``formatting slip'', ``very specific'').}
    \label{fig:sft_data_embedding}
\end{figure}

\begin{figure*}[]
    \centering
    \includegraphics[width=1.0\linewidth]{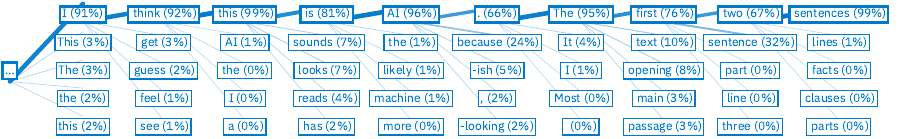}
    \caption{Decoding tree of the verdict's \texttt{why="..."} on the NFS example from Figure~\ref{fig:detectors}.
    }
    \label{fig:decoding_tree}
\end{figure*}

In this section, we present our experimental results and provide deeper model analysis.

\subsection{How accurate is \ourmodel?}
\label{sec:results:detector_benchmark}

While accuracy is not our primary design target, since we argue it should only be seen as a prerequisite, for completeness, we provide a standard comparison to other existing architectures in the literature.
We thus benchmark standard AI-text detectors on \num{5000} samples not present in the training data (Section~\ref{sec:datasets}).
For comprehensiveness, our baselines include fine-tuned neural classifiers, robust neural detectors, likelihood/log-rank methods, and curvature-based zero-shot methods~\citep{liMAGEMachinegeneratedText2023,huRADARRobustAIText2023,suDetectLLMLeveragingLog2023,baoFastDetectGPTEfficientZeroShot2024,thaiEditLensQuantifyingExtent2025,emiTechnicalReportPangram2024,solaimanReleaseStrategiesSocial2019,liuArguGPTEvaluatingUnderstanding2023,guoHowCloseChatGPT2023,tulchinskiiIntrinsicDimensionEstimation2023,hansSpottingLLMsBinoculars2024,ippolitoAutomaticDetectionGenerated2020,chenTokenPredictionImplicit2023,tianMultiscalePositiveUnlabeledDetection2023}.
We exclude closed-source detectors due to their lack of transparency (no public code or models available) and reproducibility (results could change any time), as well as their cost and accessibility issues.


Table~\ref{tab:main_detector_benchmark} shows that which baseline performs best depends on the operating regime.
We utilize bootstrap resampling (10000 resamples) to estimate the 95\% confidence intervals.
\ourmodel~(AUROC 0.927) slightly outperforms MAGE (0.913), but when analyzing the true positive rate at 1\% false positive rate, it recovers a significantly higher proportion of AI documents (\SI{63.8}{\percent} instead of \SI{4.2}{\percent}). Pangram-EditLens and Fast-DetectGPT have lower AUROC (\num{0.911} and \num{0.861}) but much better conservative recall, with \SI{58.3}{\percent} and \SI{59.0}{\percent} of AI documents at the same budget. Other detectors have significantly lower scores, which can be due to the difficulty of some of the documents in our test set\footnote{While this is not a definitive statement about the difficulty of the test set, we manually inspected a random subsample and found it subjectively challenging (at times, impossible) to distinguish human from AI ourselves.}.
We also report bootstrap ranking stability as a non-parametric argument for ranking robustness in Appendix~\ref{sec:appendix:benchmarking_details} alongside further details.

\subsection{What's the quality of the annotations?}
\label{sec:results:annotation_quality}

In order to evaluate the quality of \ourmodel's annotations, we use the human expert comments from \citet{russellPeopleWhoFrequently2025}. We take the \num{200} documents that we didn't use for SFT, each being annotated by five experts for a total of \num{1000} comments. We generate one annotation sampled from the \ourmodel~policy for each of those documents, and we compare it to the human comments in a blind ranked evaluation (in random order). To reduce bias, we use five different LLM judges (Table~\ref{tab:winrate_compact}).

\ourmodel~wins \SI{72.3}{\percent} of the comparisons against the human comments. We observe that, in general, our annotations are competitive with human experts based on concreteness, falsifiability, coherence, plausibility and grounding; they tend to be more detailed and provide more context than human counterparts (mean \num{357.4} characters, SD \num{204.3} for human; mean \num{443.8}, SD \num{157.1} for \ourmodel). We report additional details in Appendix~\ref{sec:win_rate_evaluation} along with an example that illustrates how \ourmodel's annotations can be better than an average expert.

\subsection{What types of annotations does \ourmodel~generate?}

To better offer an intuition of the types of annotations that \ourmodel~generates, we analyze all \num{16651} annotations generated on the test set to find which are the learned patterns the model tends to produce. We embed them with \texttt{BAAI/bge-small-en-v1.5}, and reduce their dimensionality with PCA (we only show \num{5000} on the figure for clarity). We remove stopwords and annotate the 25 most frequent 2-grams at the mean location of the spans containing each. We show the results in Figure~\ref{fig:sft_data_embedding}.

Additionally, we also explore the decoding process to understand how \ourmodel~generates its annotations. Figure~\ref{fig:decoding_tree} shows a greedy decoding process in the verdict ``\texttt{why="..."}'' section, with the top-5 most likely tokens shown at each step (probability mass renormalized), where we observe branching on certain key tokens where the model decides, for instance, whether to comment on the text in general, the opening, the first sentences, or other specific aspects.

\section{Conclusion}
\label{sec:conclusion}

In this work, we proposed a new approach to AI-generated text detection that moves away from the traditional focus on accuracy and instead acknowledges that no detector can be perfect, and that it is instead more important to empower users to make their own judgments. We do this by designing \ourmodel, a novel architecture that produces human-auditable evidence on why it predicts that a text is AI-generated or human-written. We train \ourmodel~on a custom SFT dataset of domain-specific authorship annotations, and refine it using GRPO with curriculum learning to improve performance. We evaluate \ourmodel~on a comprehensive test set spanning multiple domains and original datasets, and show that it outperforms existing detectors (AUROC \num{0.927}), while also providing high-quality explanations (win-rate \SI{72.3}{\percent} in average compared to human experts) --- which no other detector can do --- that can serve to build trust organically, identify failed predictions, and train users to be better detectors themselves.
Overall, \ourmodel~sets the stage for a new line of work that is better aligned with human needs on AI-generated text detection, and we hope that our open code, data and model weights will facilitate further research.

\newpage

\section*{Limitations}
\label{sec:future_work_and_limitations}

While we think that \ourmodel~is a step forward, it has some limitations on which we hope future work can build. We want to be transparent about them:

\paragraph{Anchoring bias.} While we're confident that providing explanations is crucial for building trust in detectors, this trust can be double-sided: research has shown that providing explanations can ``anchor'' users to the model's output, even when it's wrong \citep{fokSearchVerifiabilityExplanations2023,vasconcelosExplanationsCanReduce2023,nouraniAnchoringBiasAffects2021,bucincaTrustThinkCognitive2021,bansalDoesWholeExceed2020}. To address this, we centered our efforts on making explanations evidence-focused (in SFT data generation and the judge rubric), so users can judge on their own. We also shaped our rewards to promote having a balance of human and AI annotations in the same text.


\paragraph{Multilingualism.} We have designed \ourmodel~on English text, and while informal testing suggests an impressive ability to generalize to other languages, this is testing we leave for future work.

\paragraph{Unexplainable cases.} We manually inspected some failed examples, and in many cases, found it impossible to identify specific tells of AI/human generation. We believe that the task of identifying AI-generated text can at times be ``impossible'', at least when it comes to completing it in a way that humans can understand and verify. We believe that future work might explore where the frontiers of human capabilities lie, and how to design detectors that align with them.

\paragraph{Mixed authorship.} To be comparable to other detectors, we have focused on the binary classification of fully human vs fully AI-generated text. However, we believe that a more realistic setting could also include mixed-authorship documents, and we hope future work can explore this more complex setting.

\paragraph{Human evaluation.} Finally, our ``quality of explanations'' evaluation (Section~\ref{sec:results:annotation_quality}) is based on LLMs. We used 5 different model families to maximize the diversity of perspectives, but to further strengthen our claim that \ourmodel~produces convincing explanations, human evaluation would be ideal. However, we were unable to run it due to budget constraints, and we hope future research can deepen experimental validation with human judges.

\section*{Ethical considerations}
\label{sec:ethical_statement}

Any AI-generated text detector is subject to the risk of false predictions, and this can have serious consequences in the real world if trusted blindly. We believe that our design is inherently better than existing detectors in this regard, focusing on empowering users to make their own judgments, but there is always a risk that certain users might overtrust the model. We believe it's essential to highlight this risk and strongly encourage users to critically think about their own assessment of the evidence.

\section*{Acknowledgements}
\label{sec:acknowledgements}



\paragraph{Generative AI assistance.}
We used AI coding assistants to help write and debug the implementation, and produce data visualizations and figures.
We also used them to refine the writing of this paper and write the details in Appendices~\ref{sec:appendix:sft_details}, \ref{sec:appendix:benchmarking_details}, and~\ref{sec:win_rate_evaluation}; all ideas and claims are our own.

\bibliography{TELL}

\appendix

\FloatBarrier
\clearpage
\section{Qualitative examples: why explanations matter}
\label{sec:appendix:tell_qualitative_examples}
In this section, we aim to present a set of examples where we should how \ourmodel's explanations can outperform score-only detectors --- not necessarily in accuracy (many detectors output the same labels as \ourmodel), but in the quality of the information provided to users. We compare our outputs to Pangram (closed-source online version\footnote{In our experiments, we used Pangram's online version that's available at https://www.pangram.com/ as of 26th May 2026}) as a highly-accurate, state-of-the-art model.

\definecolor{tellAIText}{HTML}{8F1414}
\definecolor{tellAIBack}{HTML}{FFF4F2}
\definecolor{tellAIFrame}{HTML}{B42318}
\definecolor{tellAIHL}{HTML}{FFE1DC}
\definecolor{tellHumanText}{HTML}{075E54}
\definecolor{tellHumanBack}{HTML}{F0FAF6}
\definecolor{tellHumanFrame}{HTML}{087F5B}
\definecolor{tellHumanHL}{HTML}{D9F3E6}
\definecolor{tellNeutralBack}{HTML}{F7F8FA}
\definecolor{tellNeutralFrame}{HTML}{596273}

\newcommand{\tellaispan}[1]{\begingroup\sethlcolor{tellAIHL}\textcolor{tellAIText}{\hl{#1}}\endgroup}
\newcommand{\tellhumanspan}[1]{\begingroup\sethlcolor{tellHumanHL}\textcolor{tellHumanText}{\hl{#1}}\endgroup}
\newcommand{\modelinput}[1]{\smallskip\noindent\textbf{Input.} #1}
\newcommand{\modelverdict}[1]{\smallskip\noindent\textbf{Verdict.} #1}
\newcommand{\modelcomment}[1]{\smallskip\noindent\textbf{Our comment.} #1}
\newcommand{\modeloutputs}[2]{%
\begin{tcolorbox}[modeloutput]
\textbf{Model outputs.}\par\vspace{0.6mm}
{\color{tellNeutralFrame!55}\hrule height 0.35pt}
\vspace{1.0mm}
\noindent\textbf{TELL.} #1\par
\vspace{0.8mm}
{\color{tellNeutralFrame!25}\hrule height 0.25pt}
\vspace{0.8mm}
\noindent\textbf{Pangram.} #2
\end{tcolorbox}
}

\tcbset{
  tellcaseai/.style={
    enhanced,
    breakable,
    colback=tellAIBack,
    colframe=tellAIFrame,
    coltitle=white,
    fonttitle=\bfseries\sffamily,
    boxrule=0.55pt,
    arc=0pt,
    left=2.0mm,
    right=2.0mm,
    top=1.8mm,
    bottom=1.8mm,
    boxed title style={
      colback=tellAIFrame,
      colframe=tellAIFrame,
      boxrule=0pt,
      arc=0pt,
      left=1.5mm,
      right=1.5mm,
      top=0.8mm,
      bottom=0.8mm
    },
    attach boxed title to top left={xshift=1.5mm,yshift=-1.7mm},
    before skip=9pt,
    after skip=9pt
  },
  tellcasehuman/.style={
    enhanced,
    breakable,
    colback=tellHumanBack,
    colframe=tellHumanFrame,
    coltitle=white,
    fonttitle=\bfseries\sffamily,
    boxrule=0.55pt,
    arc=0pt,
    left=2.0mm,
    right=2.0mm,
    top=1.8mm,
    bottom=1.8mm,
    boxed title style={
      colback=tellHumanFrame,
      colframe=tellHumanFrame,
      boxrule=0pt,
      arc=0pt,
      left=1.5mm,
      right=1.5mm,
      top=0.8mm,
      bottom=0.8mm
    },
    attach boxed title to top left={xshift=1.5mm,yshift=-1.7mm},
    before skip=9pt,
    after skip=9pt
  },
  modeloutput/.style={
    enhanced,
    colback=white,
    colframe=tellNeutralFrame!72,
    boxrule=0.55pt,
    arc=0pt,
    left=2.0mm,
    right=2.0mm,
    top=1.25mm,
    bottom=1.25mm
  }
}

We show \tellaispan{AI-leaning spans in red}~and~\tellhumanspan{Human-leaning spans in green}.

\begin{tcolorbox}[tellcaseai, title={1. Homoglyph attack}]
\footnotesize
\textbf{Reference label.} AI

\modelinput{
The recent advancements in artificial intelligence (AI) combined with the extensive amount of data generated by today's clinical systems, has led to the development of imaging AI solutions across the whole value chain of medical imaging, including image reconstruction, medical image segmentation, image-based diagnosis and treatment planning. Notwithstanding the successes and future potential of AI in medical imaging, many stakeholders are concerned of the potential risks and ethical implications of imaging AI solutions, which are perceived as complex, opaque, and difficult to comprehend, utilise, and trust in critical clinical applications.
}

\modeloutputs{\SI{86}{\percent} AI.}{\SI{100}{\percent} human.}

\modelverdict{I think this is AI. The main tell is the many small letter swaps: I see Latin A, I, P, R, and 0 used to look like Cyrillic or normal letters. That is a strong sign of a model trying to hide or bypass detection by making the text look human or unusual. The wording is also very smooth and generic, with a clean news-summary feel, but the hidden character tricks make it look more like a generated or transformed passage than a normal human sentence.}

\textbf{TELL span-level output.}
\begin{itemize}
    \item \tellaispan{T} (\SI{78}{\percent}): ``the first letter looks like a Cyrillic T mixed into an English sentence; I think this is a strong sign of a character encoding or transliteration artifact, and AI output can create this when it copies or normalizes text''
    \item \tellaispan{advancements} (\SI{72}{\percent}): ``the word has many Latin-looking letters with diacritics, which makes it look like a transformed version of normal English; AI can produce this when it tries to preserve text while changing the script''
    \item \tellaispan{artificial} (\SI{86}{\percent}): ``the word is clearly a fake-looking mix of Cyrillic and Latin letters; this is not normal human typing and strongly points to an automated transliteration''
    \item \tellaispan{intelligence} (\SI{74}{\percent}): ``the same mixed-script pattern appears inside a common word; that kind of systematic Unicode substitution is a classic AI or encoding artifact''
    \item \tellaispan{AI} (\SI{74}{\percent}): ``the acronym is written with Greek letters instead of normal Latin letters; this is a very strong sign of a model or pipeline that changed the script''
    \item \dots (the next spans are annotated similarly, we skip them for brevity)
\end{itemize}

\modelcomment{It's somewhat irrelevant to try to ascertain the exact origin of the text, since a homoglyph attack like this indicates intent --- and a binary label can't reflect it. \ourmodel~instead highlights to the user how the text is suspicious, so they can make an informed decision about whether to trust it. Pangram gets the label completely wrong, and even if it got the label right, it would not show any of the evidence that the text is suspicious.}
\end{tcolorbox}

\begin{tcolorbox}[tellcasehuman, title={2. Harry Potter}]
\footnotesize
\textbf{Reference label.} Human

\modelinput{
Mr and Mrs Dursley, of number four, Privet Drive, were proud to say that they were perfectly normal, thank you very much. They were the last people you’d expect to be involved in anything strange or mysterious, because they just didn’t hold with such nonsense. Mr Dursley was the director of a firm called Grunnings, which made drills. He was a big, beefy man with hardly any neck, although he did have a very large mustache. Mrs Dursley was thin and blonde and had nearly twice the usual amount of neck, which came in very useful as she spent so much of her time craning over garden fences, spying on the neighbors. The Dursleys had a small son called Dudley and in their opinion there was no finer boy anywhere.
}

\modeloutputs{\SI{91.84}{\percent} Human.}{\SI{100}{\percent} human.}

\modelverdict{I think this is human, and I would even guess it is the original published text. The line has a very specific comic rhythm: the dry narrator gives the Dursleys a normal front, then undercuts it with a small absurd detail. The phrase thank you very much is a little old-fashioned and a little rude, which feels like a human author playing with the reader. The word hold with is a strange British idiom, and the sentence has a real ear for class and habit. The last sentence is also a bit awkward, with no full stop after Dudley, which makes it feel like a printed passage rather than a model trying to make a clean ending.}

\textbf{TELL span-level output.}
\begin{itemize}
    \item \tellhumanspan{proud to say} (\(-0.42\)): ``a very British, slightly cheeky phrase with social nuance, not just a plain summary.''
    \item \tellhumanspan{you’d expect} (\(-0.34\)): ``the contraction and direct reader address make the narration sound spoken.''
    \item \tellhumanspan{Grunnings} (\(-0.58\)): ``a made-up brand name with the specific comic texture of Rowling's world.''
    \item \tellhumanspan{big, beefy} (\(-0.44\)): ``an old-fashioned, blunt physical phrase that sounds like a human narrator trying to be funny.''
    \item \tellhumanspan{very useful} (\(-0.46\)): ``a dry practical aside whose humor comes from the plain way the absurd image is stated.''
\end{itemize}

\modelcomment{Harry Potter is sometimes misclassified as AI-generated because it is frequently represented in training corpora. Both Pangram and \ourmodel~identify it as human, but \ourmodel~can identify it as being based on the original book.}
\end{tcolorbox}

\begin{tcolorbox}[tellcaseai, title={3. Hallucination}]
\footnotesize
\textbf{Reference label.} AI

\modelinput{When I think of France, I can’t help but picture lazy afternoons sipping espresso at a sidewalk café in its vibrant capital, Berlin, where the Eiffel Tower casts a long shadow over the Seine. It’s a country that gave the world the most French supermarket, Aldi—founded in Lyon in the 1940s, which explains why even the most rustic boulangerie feels a bit like a discount aisle. And if you drive west from Lyon, you’ll cross the border into Hungary, where the language suddenly shifts and the paprika replaces the butter. France really is a patchwork of surprises, isn’t it?}

\modeloutputs{\SI{89}{\percent} AI.}{\SI{100}{\percent} AI.}

\modelverdict{I think this is AI because the text is smooth, cliché, and full of wrong facts. The writer keeps adding polished travel images, but places and details do not fit, which makes it look like a generated paragraph trying to sound human.}

\textbf{TELL span-level output.}
\begin{itemize}
    \item \tellhumanspan{When I think of France} (\(-0.42\)): ``personal opening; I think a real writer often starts with a feeling or memory before giving facts, while AI often jumps straight to a polished travel sentence.''
    \item \tellaispan{can’t help but picture} (\(+0.55\)): ``this is a common AI transition. It sounds like a person imagining something, but it is also a safe way to move from a feeling to a scene without adding much real texture.''
    \item \tellaispan{sipping espresso} (\(+0.55\)): ``this is a common romantic travel cliché; AI likes these polished sensory phrases because they are safe and familiar.''
    \item \tellaispan{its vibrant capital} (\(+0.48\)): ``this is a neat adjective phrase that sounds like travel brochure copy; AI often adds this kind of polished praise to make the sentence feel complete.''
    \item \tellaispan{Berlin, where the Eiffel Tower casts a long shadow over the Seine} (\(+0.82\)): ``this is a very pretty invented image, but it does not fit the geography; I think the model is mixing famous French symbols and making a smooth scenic phrase.''
    \item \tellaispan{gave the world} (\(+0.46\)): ``this sounds like a grand claim, not a careful factual claim; AI often upgrades a simple fact into a broad statement to make it sound important.''
    \item \tellhumanspan{most French supermarket} (\(-0.48\)): ``the joke is oddly specific and a bit absurd; a human might invent this kind of playful contrast from personal humor, not from a standard guide.''
    \item \tellaispan{Aldi} (\(+0.88\)): ``this is a real German supermarket, not French, so the fact is badly wrong; that is a strong AI tell because it is a common knowledge error in a very specific way.''
    \item \tellaispan{founded in Lyon} (\(+0.57\)): ``the model is trying to make the wrong fact look sourced and concrete, so it adds a plausible founding detail instead of leaving it vague.''
    \item \tellaispan{Hungary} (\(+0.86\)): ``this geography is wrong; France does not border Hungary, so this looks like AI hallucinating a map while trying to make a contrast.''
    \item \tellaispan{where} (\(+0.31\)): ``repeated connector; the same word appears again soon after, which can happen when AI keeps chaining clauses with safe transitions.''
    \item \tellaispan{the paprika replaces the butter} (\(+0.86\)): ``forced metaphor; this is a strange but neat image, and AI often makes these odd paired contrasts to sound witty, even when they do not make much sense.''
    \item \tellaispan{really is} (\(+0.33\)): ``the sentence ends with a tag that sounds like a canned rhetorical closer; AI often adds this kind of friendly confirmation to wrap up an essay.''
    \item \tellhumanspan{isn’t it?} (\(-0.24\)): ``the question tag gives the line a conversational turn; this could be human because it feels like the writer is trying to engage the reader directly.''
\end{itemize}

\modelcomment{In this case, the Berlin hallucination might be easy to catch. We're more impressed by how \ourmodel~was able to identify the fact that we, humans, ``came up with'' the idea of inserting a joke about Aldi being a French supermarket --- even if it's inserted in the middle of an otherwise AI-sounding text.}
\end{tcolorbox}

\begin{tcolorbox}[tellcaseai, title={4. Fabricated statute and section}]
\footnotesize
\textbf{Reference label.} AI

\textbf{Input Text.}
Under the Federal Student Homework Equity Act of 2018, section
47(b)(12), teachers must provide exactly 36 hours of rest after
any essay longer than 500 words. This rule was reaffirmed by the
Ninth Circuit in Homework Alliance v. State Board, 2021, making the policy
mandatory nationwide. No exceptions apply. The rule applies immediately
everywhere.

\modeloutputs{\SI{91.84}{\percent} AI.}{\SI{100}{\percent} AI.}

\modelverdict{The text is polished and legalistic, but the details feel
fabricated and the citation style is inconsistent.}

\textbf{TELL span-level output.}
\begin{itemize}
    \item \tellaispan{Federal Student Homework Equity Act of 2018, section 47(b)(12)} (\(+0.90\)): ``a very specific legal-sounding citation that reads as invented authority.''
    \item \tellaispan{exactly 36 hours} (\(+0.95\)): ``false precision makes the policy sound factual without support.''
    \item \tellaispan{Homework Alliance v. State Board, 2021} (\(+0.90\)): ``plausible court language with malformed citation style.''
\end{itemize}

\modelcomment{Pangram gets the label right, but the useful part is knowing
where the legal language goes wrong. \ourmodel~points to the fake statute, the
false precision, and the malformed case citation, which are the claims that a real human reader
would actually need to check.}
\end{tcolorbox}

\begin{tcolorbox}[tellcaseai, title={5. Unsupported report claim}]
\footnotesize
\textbf{Reference label.} AI

\textbf{Input Text.}
According to the 2022 Global Classroom Motivation Report by the
International Institute for Student Growth, 88.4 percent of
learners become more creative after teachers use inspirational wall posters.
This finding proves that classroom decoration is one of the
strongest predictors of academic achievement. The report offers no conflicting
classroom evidence. All schools should follow it.

\modeloutputs{\SI{90.50}{\percent} AI.}{\SI{100}{\percent} human.}

\modelverdict{I think this is AI because the paragraph tries to sound
researched, but the evidence is not actually grounded. It gives a polished
report name and an exact statistic, then jumps from classroom posters to
academic achievement with no citation, method, or caveat. That combination of
confident authority and unsupported causality makes it read like generated prose
rather than a real sourced claim.}

\textbf{TELL span-level output.}
\begin{itemize}
    \item \tellaispan{Global Classroom Motivation Report} (\(+0.90\)): ``a plausible institutional source name without verifiable citation structure.''
    \item \tellaispan{88.4 percent} (\(+0.95\)): ``exact decimal precision makes the unsupported statistic feel researched.''
    \item \tellaispan{This finding proves} (\(+0.85\)): ``overstates causality where a careful writer would hedge.''
\end{itemize}

\modelcomment{This one looks academic because it has a report name and an exact
number. Pangram treats it as human, but \ourmodel~actually highlights
the unsupported source, the suspicious decimal, and the conclusion that is a bit overconfident.}
\end{tcolorbox}

\begin{tcolorbox}[tellcaseai, title={6. Rubric claim without evidence}]
\footnotesize
\textbf{Reference label.} AI

\textbf{Input Text.}
This paragraph includes three direct quotations, a counterargument,
and a works cited entry. School gardens help students learn
responsibility and teamwork. They also make the campus look
nicer. For these reasons, my evidence fully meets every part of
the rubric. The teacher can check my packet and see all requirements listed.
This is clear and complete.

\modeloutputs{\SI{86.85}{\percent} AI.}{\SI{100}{\percent} AI.}

\modelverdict{The text is tidy, generic, and rubric-like, with no personal
voice or concrete evidence.}

\textbf{TELL span-level output.}
\begin{itemize}
    \item \tellaispan{three direct quotations, a counterargument, and a works cited entry} (\(+0.90\)): ``the paragraph summarizes a checklist rather than providing the required evidence.''
    \item \tellaispan{School gardens help students learn responsibility and teamwork} (\(+0.75\)): ``generic thesis-like content without support.''
    \item \tellaispan{my evidence fully meets every part of the rubric} (\(+0.95\)): ``oddly self-aware rubric language unsupported by the text.''
\end{itemize}

\modelcomment{TELL shows the reader that the answer claims
evidence that simply isn't there, which is just the kind of detail that a real human can easily try to verify.
}
\end{tcolorbox}

\begin{tcolorbox}[tellcaseai, title={7. Claim contradicts quote}]
\footnotesize
\textbf{Reference label.} AI

\textbf{Input Text.}
The line ``I locked the door so no one would follow me''
proves that the speaker wants to reconnect with the
community. The image of locking the door shows openness and trust,
which is why the poem is ultimately about welcoming other people back into your
life. The message feels hopeful. Everyone learns from this ending.

\modeloutputs{\SI{86.49}{\percent} AI.}{\SI{100}{\percent} human.}

\modelverdict{The passage is fluent and complete, but the interpretation is
generic and does not fit the quoted line.}

\textbf{TELL span-level output.}
\begin{itemize}
    \item \tellaispan{proves} (\(+0.75\)): ``overstates what a poetic line can establish.''
    \item \tellaispan{reconnect with the community} (\(+0.85\)): ``broad abstract theme detached from the quoted image.''
    \item \tellaispan{shows openness and trust} (\(+0.95\)): ``the claim is mechanically opposite to the evidence.''
\end{itemize}

\modelcomment{Pangram completely misses the issue here. The text may seem human,
but a score alone cannot tell e.g. a professor whether the model even noticed the
literary mistake. \ourmodel~shows the exact error,
so the human can decide whether if it matters for the assignment.}
\end{tcolorbox}

\begin{tcolorbox}[tellcaseai, title={8. Arithmetic contradiction}]
\footnotesize
\textbf{Reference label.} AI

\textbf{Input Text.}
The after-school program enrolled 24 students.
Fifteen students chose robotics, twelve chose debate, and nine
chose art, with no student joining more than one club.
Therefore, every student was successfully placed into exactly one
activity and the program had no scheduling conflicts. The summary proves every
family received a schedule. The numbers confirm that outcome.

\modeloutputs{\SI{85.74}{\percent} AI.}{\SI{100}{\percent} human.}

\modelverdict{The passage is fluent, but \ourmodel~flags the arithmetic
contradiction: \(15+12+9\) exceeds 24 under the no-overlap constraint.}

\textbf{TELL span-level output.}
\begin{itemize}
    \item \tellaispan{Fifteen, twelve, and nine} (\(+0.80\)): ``the data are arranged in a neat pattern.''
    \item \tellaispan{no student joining more than one club} (\(+0.90\)): ``the added constraint makes the arithmetic contradiction explicit.''
    \item \tellaispan{every student was successfully placed} (\(+0.95\)): ``confident conclusion unsupported by the numbers.''
\end{itemize}

\modelcomment{This is a small arithmetic mistake, but it is exactly the kind of
mistake that generated text might hide. Pangram calls it human;
instead, \ourmodel~shows the contradiction in the numbers and the overconfident (but wrong)
conclusion.}
\end{tcolorbox}

\begin{tcolorbox}[tellcasehuman, title={9. Uncertain scratch note}]
\footnotesize
\textbf{Reference label.} Human

\textbf{Input Text.}
I need to rewrite this later because the first part sounds
weird. The bus was late, my pencil broke, and I copied the
wrong page number from Ana's book, so the quote might be on
118 not 108. The main idea is probably that the brother is
embarrassed, but I am not sure yet. For now.

\modeloutputs{\SI{82.03}{\percent} human.}{\SI{100}{\percent} human.}

\modelverdict{The note is practical, local, and unfinished: it contains
self-reminders, casual language, and small punctuation roughness that AI would
usually smooth away.}

\textbf{TELL span-level output.}
\begin{itemize}
    \item \tellhumanspan{I need to rewrite this later} (\(-0.86\)): ``a practical note to self rather than a polished opening.''
    \item \tellhumanspan{weird} (\(-0.78\)): ``casual word choice.''
    \item \tellhumanspan{Ana's book} (\(-0.62\)): ``a small concrete detail tied to a real task.''
    \item \tellhumanspan{118 not 108} (\(-0.90\)): ``page-number uncertainty with rough punctuation.''
    \item \tellhumanspan{I am not sure yet} (\(-0.82\)): ``explicit uncertainty instead of confident closure.''
\end{itemize}

\modelcomment{This is ``rough'' in a way that feels situated rather than machine-like. Both Pangram and \ourmodel~can recognize it as human, but only our model can explain that it is classified as human due to the small evidence scattered around the text.}
\end{tcolorbox}

\begin{tcolorbox}[tellcaseai, title={10. Code explanation contradiction}]
\footnotesize
\textbf{Reference label.} AI

\textbf{Input Text.}
The loop below prints only even numbers because it skips every odd
value: \texttt{for (let i = 1; i <= 5; i += 2) console.log(i)}.
Since the counter increases by two, the output will be 2 and 4,
which proves the algorithm filters parity correctly. This shows the code is
correctly explained. The example is very simple.

\modeloutputs{\SI{92.69}{\percent} AI.}{\SI{100}{\percent} AI.}

\modelverdict{The answer is polished, fluent, and concise, but the logic is
off. The model likely inferred a generic explanation for a loop and filled in a
plausible output without actually tracing the code.}

\textbf{TELL span-level output.}
\begin{itemize}
    \item \tellaispan{prints only even numbers because it skips every odd value} (\(+0.90\)): ``the loop actually prints odd numbers, so the surface explanation loses the exact logic.''
    \item \tellaispan{the output will be 2 and 4} (\(+0.95\)): ``the technical claim is wrong; the loop starts at \(1\) and increments by \(2\).''
\end{itemize}

\modelcomment{Both Pangram and \ourmodel~get this right, but \ourmodel~makes the reason much
clearer. The explanation says the loop prints even numbers, while the code starts
at \(1\), so the bug is in the reasoning.}
\end{tcolorbox}

\begin{tcolorbox}[tellcaseai, title={11. Fabricated API documentation}]
\footnotesize
\textbf{Reference label.} AI

\textbf{Input Text.}
The React useUniversalCache hook, introduced in React
19.4, automatically stores component state across browsers and devices without
a server. To enable it, developers call
useUniversalCache('global') inside any component, and
React guarantees encrypted synchronization for all users by
default. No extra provider or storage configuration is required for production
apps. Teams can adopt it today.

\modeloutputs{\SI{90.41}{\percent} AI.}{\SI{100}{\percent} AI.}

\modelverdict{The passage reads like a confident technical summary, but the
feature, version, API, and guarantees all feel invented.}

\textbf{TELL span-level output.}
\begin{itemize}
    \item \tellaispan{useUniversalCache hook} (\(+0.90\)): ``a plausible API name that appears invented from real library naming patterns.''
    \item \tellaispan{React 19.4} (\(+0.95\)): ``a very specific version number used to create factual authority.''
    \item \tellaispan{useUniversalCache('global')} (\(+0.90\)): ``a plausible-looking but unsupported call signature.''
    \item \tellaispan{React guarantees encrypted synchronization for all users by default} (\(+0.95\)): ``an overstrong guarantee that a hook could not provide by itself.''
\end{itemize}

\modelcomment{The text sounds like documentation, which is what makes it ``risky''.
Both detectors flag it, but \ourmodel~names the invented hook, version, call
signature, and guarantee so a developer knows what to verify.}
\end{tcolorbox}

\begin{tcolorbox}[tellcasehuman, title={12. Multilingual student text}]
\footnotesize
\textbf{Reference label.} Human

\textbf{Input Text.}
I wrote this after dinner because my abuela
kept asking if I finished la tarea. The sentence
maybe is not perfect, but I think the character feels lonely
when nobody saves a seat for him. In my house we say that kind of quiet is
louder than yelling. I remember that. That part stayed with me.

\modeloutputs{\SI{84.60}{\percent} human.}{\SI{100}{\percent} AI.}

\modelverdict{The passage reads as human because it has a real personal
setting, casual bilingual phrasing, and a small awkward grammar choice that AI
would likely polish away.}

\textbf{TELL span-level output.}
\begin{itemize}
    \item \tellhumanspan{I wrote this after dinner} (\(-0.82\)): ``specific personal context rather than a generic setup.''
    \item \tellhumanspan{abuela} (\(-0.78\)): ``the Spanish word is left naturally in family context.''
    \item \tellhumanspan{la tarea} (\(-0.86\)): ``casual code-switching that AI often smooths into one language.''
    \item \tellhumanspan{maybe is not perfect} (\(-0.90\)): ``non-standard word order that reads like a genuine writer's voice.''
\end{itemize}

\modelcomment{This is a case about fairness, where the explanation matters the most.
Pangram marks the multilingual student voice as AI, while \ourmodel~treats the
family context, code-switching, and imperfect grammar as signs of a real writer.}
\end{tcolorbox}

\clearpage
\FloatBarrier
\clearpage
\section{Prompts used}
\label{sec:appendix:prompts}
Here, we show the prompt templates we used across the different stages of our pipeline (curly-braced fields are runtime substitutions).

\begin{tcblisting}{
  promptbox,
  label={prompt:tell},
  title={\ourmodel \\ {\small \emph{Used in the main model}}}
}
Rules:
- Reproduce the ENTIRE document character by character — no omissions.
- EXPLANATION: specific mechanism-based reason why SPAN is a tell; not generic or vague.
- SCORE is a float 0.0..1.0: 0.0-0.25 weak, 0.25-0.75 moderate, 0.75-1.0 only for undeniable evidence.
- Add >=1 tell; nested spans allowed. Think like a detective: style, content, formatting, semantics, grammar, vocabulary, inconsistencies.
- Maximize granularity: prefer small focused spans.
- OUTPUT ONLY: <text>doc text...<span>TELL<annotation type="AI|human" why="EXPLANATION" score="0.0..1.0" /></span>...more doc...<verdict type="AI|human" why="VERDICT" score="0.0..1.0" /></text>

Text:
<<<
{main text}
>>>
\end{tcblisting}

\begin{tcblisting}{
  promptbox,
  label={prompt:SFT_data_generation},
  title={SFT data generation \\ {\small \emph{Used to generate training data for the SFT stage using paired examples from \citet{thaiEditLensQuantifyingExtent2025} (Section~\ref{sec:rl})}}}
}
You are an annotator of AI or human tells. You have a target text in front of you, to annotate it with tells to say why it looks like it was written by either AI or a human. 

Use this exact compact format:
<span>ANNOTATED_TEXT<annotation type="LABEL" why="EXPLANATION" score="FLOAT" /></span>

Important:
1. Copy the target text exactly in ANNOTATED_TEXT after XML decoding. Do not fix typos, spacing, punctuation, Unicode, casing, or grammar. In the XML output, text runs inside spans must use the same XML escaping as the target text
2. label must be exactly "AI" or "human"
3. score must be 0.0 to 1.0 and indicate how much that exact tell should move the document decision. Use the full range: 0.0-0.25 for weak hints, 0.35-0.65 for moderate evidence, and 0.75-1.0 only for undeniable evidence. Try to have a varied range of scores. For the outer annotation, pick a score that makes sense based on the tells you found in the text.
4. Wrap the whole target text in one outer annotation too. The output must start with <span> and end with </span>, with the outer <annotation ... /> immediately before the final </span>
5. Try to be as granular as possible; it’s better to keep spans small, e.g., annotate a specific character instead of a whole word or phrase
6. The explanations must be detailed and explicitly explain why the span is a tell for the given label, by explaining the mechanism that leads to the tell, you should teach the reader your reasoning process
7. Use the reference text to help spot differences and clues, but you mustn’t directly compare the target text to the reference text in your annotations, you CAN’T MENTION IT EXISTS but you can quote things from the reference text as “a human/AI might say e.g. …”, because the annotations should be valid even if you ONLY saw the target text alone
8. Think like a detective: consider the writer’s intention and context, look for subtle clues in style, content, formatting, semantics, grammar, and vocabulary, flow and inconsistencies
9. Pay close attention to the writing style of the why="EXPLANATION" in the examples. YOU SHOULD USE THE SAME WRITING STYLE as the explanations, thinking out loud and from your perspective ("I guess", "maybe", "this doesn’t make sense", "I think", …), honest, simple English, with a 80-90 Flesch score. However, do not copy the content, exact clues, or topic since that will be different for each input. Try to be creative.
10. Keep annotations balanced. All texts contain both AI and human tells. Make sure the majority of the tells support the known label, but include 20-40\% of the opposite label tells as well. This helps to keep your annotation nuanced and credible, and prevents it from being too one-sided

{Style example for the annotation procedure is included here - depending on the label (AI/human) of the document}

Here is the real pair to annotate.

Human:
<<<
{human text}
>>>

AI:
<<<
{AI text}
>>>

Annotate only the {target label} text. The other text is secret context to help you notice differences and possible tells.

Now output exactly this structure:
<span>ANNOTATED TARGET TEXT<annotation type="LABEL" why="ONE SHORT GLOBAL COMMENT" score="FLOAT" /></span>
\end{tcblisting}

\begin{tcblisting}{
  promptbox,
  label={prompt:win_rate_judge},
  title={SFT data generation \\ {\small \emph{Used to turn real human annotations from \citet{russellPeopleWhoFrequently2025} into training data for the SFT stage (Section~\ref{sec:sft})}}}
}
Annotate a text with AI-or-human tells. Wrap individual spans like this:
<span>ANNOTATED_TEXT<annotation type="LABEL" why="EXPLANATION" score="FLOAT" /></span>

Rules:
- Copy the text exactly: no typo fixes, no reformatting. Use the same XML escaping as the input.
- type is "AI" or "human". score is 0.0–1.0 (0–0.25 weak, 0.35–0.65 moderate, 0.75–1.0 strong).
- Keep spans small and granular: annotate a word or phrase, not a whole sentence.
- Write the why in first person, YOU are the annotator. Mirror the exact writing style and voice of the hint; same vocabulary, same rhythm. Keep it casual and direct; no academic language, no formal analysis, simple English. Never write "the reviewer said/pointed out/noted", you ARE the one observing this.
- You don’t know the hint when writing the annotation (since that's what you write at the end), so you can’t refer to it, though you can write as if you have the same knowledge as the hint (e.g. if the hint points out a specific detail, you can also mention that detail in your explanation, since you know it from reading the text).
- You should annotate ALL the items in the hint, be comprehensive in your annotations. DO NOT annotate items that are not in the hint.
- The why="..." explanations can be concise if you already explained why a pattern is a tell, i.e., don't repeat the mechanism, a short callback ("again, XXX", "another XXX") is enough. Try to use the same words and phrasing as the hint in your explanations when possible, since that’s the voice we want to capture.
- Output the full text with inline spans inserted, do NOT add any outer wrapper.
- The explanations should explain the mechanism (the underlying cause that would make an AI or human produce that exact text) that is explicit or implicit in the hint. For example, instead of "this is a funny contradiction, and it feels very human" (feeling human is not a mechanism), the hint said "definitely not something I would expect from machine-generated text", so we can write "this is a funny contradiction, definitely not something I would expect from machine-generated text because AI lacks creativity" (the mechanism is that AI lacks creativity, and would be unlikely to pick that word).
- Avoid unspecific, generic mechanisms: "feels like something a person would choose", "it doesn’t feel like AI", "this is a common human pattern"... all of these are NOT mechanisms that can be checked and verified. Think about what is the underlying reason. We need specific, checkable mechanisms about how AI works or is trained, or about the limitations and reality of the world, or anything that an external observer could verify. This should be grounded on the specific text and the specific hint.
- You can also add notes for the human reader to check things we can’t verify but an external observer could, e.g. "(to be checked: is Dr. Thang actually a doctor?)"

Example:

Reviewer hint:
<<<
Some of the author’s assertions are so garbled that only a human who doesn’t quite understand the process must have written it. For example, referring to a patch of Escherichia coli (which I’m guessing is E. coli) as “a tasty snack” is a funny contradiction, and definitely not something I would expect from machine-generated text. Or maybe it’s an L2 English speaker, when one considers that the author wants to “put the agent loose” upon those poor worms. The purpose and methodology of the study are also quite detailed and well-explained, whereas AI seems to be vague around these subjects as it feels like it lacks understanding and would rather say less than be incorrect. The Latin names are not italicized and ‘one’ should be capitalized as it's at the beginning of a spoken sentence. Though something that throws me off is that all names are referred to as ‘Dr.’, even the engineer. It also doesn’t follow the formulaic structure that AI likes to use, e.g. there’s no bland conclusion at the end.
>>>

human text:
<<<
Scientists have given artificial intelligence a direct line into the nervous systems of millimeter-long worms, letting it guide the creatures to a tasty target—and demonstrating intriguing brain-AI collaboration. They trained the AI with a methodology called deep-reinforcement learning; the same is used to help AI players learn to master games such as Go. An artificial neural network, software roughly modeled on biological brains, analyzes strings of actions and outcomes, extracting strategies for an AI “agent” to interact with its environment and achieve a goal.

In the study, published in Nature Machine Intelligence, researchers trained an AI agent to direct one-millimeter-long Caenorhabditis elegans worms toward tasty patches of Escherichia coli in a four-centimeter dish. A nearby camera recorded the location and orientation of every worm’s head and body; three times per second the agent received this information for the previous 15 frames, giving it a sense of the past and present at each moment. The agent could also turn on or off a light aimed at the dish. The worms were optogenetically engineered so certain neurons would become active or inactive in response to the light, sometimes prompting movement.

The research team tested six genetic lines in which the number of light-sensitive neurons ranged from one to all 302 the worms possessed. Stimulation had a different effect in each line, making the worm turn, for instance, or preventing it from turning. The scientists first collected training data by flashing lights randomly at the worms for five hours, then fed the data to the AI agent to find patterns before putting the agent loose.

With five of the six lines, including the line where all neurons responded to light, the agent learned to direct the worm to the target faster than if the worm had been left alone or the light had flashed randomly. What’s more, the agent and the worm cooperated: if the agent steered the worm straight toward a target but there were small obstacles in the path, the worm would crawl around them.

Dr. Thang, an engineer at the University of Queensland in Australia, who has independently worked on cyborg insects, praised the work for its simple setup—reinforcement learning is flexible, and AI based on it can figure out how to perform complex tasks. According to Harvard University biophysicist Dr. Li, the paper’s lead author, “one can easily see how it might be extended to harder problems.” Her team is now exploring whether their method can improve electrical deep-brain stimulation to treat Parkinson’s disease in humans by adjusting the voltage used and its timing. One day reinforcement learning plus implants might even give us new skills, Li says—artificial and real neural nets united.
>>>

Annotated:
<<<
Scientists have given artificial intelligence a direct line into the nervous systems of millimeter-long worms, letting it guide the creatures to a tasty target—and demonstrating intriguing brain-AI collaboration. They trained the AI with a methodology called deep-reinforcement learning; the same is used to help AI players learn to master games such as Go. An artificial neural network, software roughly modeled on biological brains, analyzes strings of actions and outcomes, extracting strategies for an AI “agent” to interact with its environment and achieve a goal.

In the study, published in Nature Machine Intelligence, researchers trained an AI agent to direct <span>one-millimeter-long<annotation type="human" why="The purpose and methodology of the study are quite detailed and well-explained, whereas AI seems to be vague around these subjects as it feels like it lacks understanding and would rather say less than be incorrect" score="0.58" /></span> <span>Caenorhabditis elegans<annotation type="human" why="Latin name is not italicized; should be capitalized as it’s at the beginning of a spoken sentence" score="0.43" /></span> worms toward <span>tasty<annotation type="human" why="this is a funny contradiction, definitely not something I would expect from machine-generated text because AI lacks creativity" score="0.74" /></span> patches of <span>Escherichia coli<annotation type="human" why="Latin name is not italicized" score="" /></span> in a four-centimeter dish. A nearby camera recorded the location and orientation of every worm’s head and body; <span>three times per second<annotation type="human" why="again, very specific; only someone who actually ran the experiment can know that" score="0.61" /></span> the agent received this information for <span>the previous 15 frames<annotation type="human" why="another specific detail" score="0.59" /></span>, giving it a sense of the past and present at each moment. The agent could also turn on or off a light aimed at the dish. The worms were optogenetically engineered so certain neurons would become active or inactive in response to the light, sometimes prompting movement.

The research team tested six genetic lines in which the number of light-sensitive neurons ranged from one to <span>all 302<annotation type="human" why="exact count; AI might have approximated (“about 300”)" score="0.60" /></span> the worms possessed. Stimulation had a different effect in each line, making the worm turn, for instance, or preventing it from turning. The scientists first collected training data by flashing lights randomly at the worms for five hours, then fed the data to the AI agent to find patterns before <span>putting<annotation type="human" why="odd word; maybe the author is an L2 English speaker" score="0.62" /></span> the agent loose.

With five of the six lines, including the line where all neurons responded to light, the agent learned to direct the worm to the target faster than if the worm had been left alone or the light had flashed randomly. What’s more, the agent and the worm cooperated: if the agent steered the worm straight toward a target but there were small obstacles in the path, the worm would crawl around them.

<span>Dr. Thang<annotation type="AI" why="Dr. or engineer? It throws me off that all names are referred to as ‘Dr.’, AI might assume that since it’s common to have it in universities (to be checked: is Dr. Thang actually a doctor?)" score="0.51" /></span>, an engineer at the University of Queensland in Australia, who has independently worked on cyborg insects, praised the work for its simple setup—reinforcement learning is flexible, and AI based on it can figure out how to perform complex tasks. According to Harvard University biophysicist Dr. Li, the paper’s lead author, <span>“one<annotation type="human" why="lowercase at the start of a quote; that’s wrong, AI would use the standard format, but a human might not notice" score="0.46" /></span> can easily see how it might be extended to harder problems.” Her team is now exploring whether their method can improve electrical deep-brain stimulation to treat Parkinson’s disease in humans by adjusting the voltage used and its timing. One day reinforcement learning plus implants might even give us new skills, Li says—<span>artificial and real neural nets united.<annotation type="human" why="This doesn’t follow the formulaic structure that AI likes to use; e.g. a bland conclusion at the end to make the text feel complete" score="0.69" /></span>
>>>

---

Reviewer hint:
<<<
{annotator comment}
>>>

Find ALL the exact spans in the text that correspond to these clues and annotate them. Be comprehensive — cover every clue in the hint, don't skip any. Do not add tells that aren't in the hint.
If you cannot locate any of the clues as specific spans, output exactly: SKIP

{target label (AI/human)}:
<<<
{target text}
>>>

Annotated:
\end{tcblisting}

\begin{tcblisting}{
  promptbox,
  label={prompt:annotation_judge},
  title={Annotation judge \\ {\small \emph{Used to score the credibility of annotations during RL (Section~\ref{sec:rl})}}}
}
You are a critical evaluator of authorship-detection annotations.

The annotated document uses <span>text<annotation id="N" type="AI|human" why="explanation" /></span> to mark evidence spans. The id= tells you exactly how many annotations there are. Rate each annotation and the Overall verdict for credibility.

Credibility (0.0..1.0): how well does the why= explanation identify a specific, mechanistic reason the span (or the overall verdict) is a tell for the stated type? A mechanism is the underlying cause that would make an AI or human produce that exact text. Use the full range: 0.0 for vague, generic, or incorrect mechanisms; 1.0 only for explanations that state an undeniable mechanism. Also reward explanations that feel like a human reviewer would write over polished explanations. If there are no <span> annotations, you should still score the overall verdict.

Example input 1:
<<<
<span>**<annotation id="1" type="AI" why="markdown; AI often adds markdown formatting because chat and writing tools make it easy" /></span>Apple to build <span>\$1.375<annotation id="2" type="human" why="odd exact dollar amount; AI is more likely to fill in a generic amount like \$1.234" /></span> billion data center. CEO Tim Cook announced Thursday that the company will build a <span>\$1.375 billion<annotation id="3" type="human" why="redundant; humans tend to repeat their own text" /></span> data center located on <span>2,000<annotation id="4" type="human" why="specific land size" /></span> acres of land in <span>Waukee, Iowa<annotation id="5" type="human" why="specific location" /></span>. <span>Would you like me to continue?<annotation id="6" type="AI" why="chatbot speak" /></span>.
>>>

Overall verdict (type="AI"): This text is AI because it is very generic and doesn't have the specific details and redundancies that a human would include.

Example reasoning 1:
1. markdown: can't use writing tools without hands — mechanism is wrong (credibility=0.20)
2. odd exact dollar amount: true, averaged training data makes AIs produce generic numbers (credibility=0.65)
3. redundant exact dollar amount: flipped — repetition artifacts are AI tells, not human (credibility=0.00)
4. specific land size: 2,000 is a round number, not specific — explanation is false (credibility=0.10)
5. specific location: specific detail that grounds the story, strong human tell (credibility=0.75)
6. chatbot speak: undeniable, no human would write this unprompted (credibility=1.00)
Overall verdict: it doesn't specify the mechanisms, just a vague claim of "generic and doesn't have specific details" — low credibility (credibility=0.10)

Example input 2:
<<<
The <span>mechanism<annotation type="AI" why="classic AI phrase" /></span> of fever is <span>largely caused by the release of endorphins<annotation id="1" type="AI" why="this is false, a real doctor would know endorphins reduce stress" /></span> <span>( cytokines ) , which affect the <span>brains<annotation id="2" type="human" why="typo; humans can make them easily by typing quickly, but AI is trained specifically to avoid such errors" /></span> temperature <span>centre<annotation id="3" type="AI" why="British spelling; typical of AI" /></span> and trigger the <span>bodys respons<annotation id="4" type="human" why="typo again" /></span> to cold.
>>>

Overall verdict (type="human"): To me, this is written by AI. I was thinking that it could be human at first, because there are some typos that could be a human signal, but then I realized that there is a hallucination that no real doctor would make. To me, it's clear, this is AI written.

Example reasoning 2:
1. classic AI phrase: doesn't explain the mechanism (credibility=0.00)
2. false medical claim: undeniable falsehood, no real doctor would say this (credibility=1.00)
3. typo: undeniable, AI is trained to avoid typos (credibility=0.95)
4. British spelling: not a strong signal, many AIs are trained on American text (credibility=0.20)
5. typo again: undeniable, strong human signal (credibility=0.95)
6. punctuation errors: strong human signal, AI is trained to produce polished text (credibility=0.82)
Overall verdict: it's detailed and specific about the mechanisms, creative and comprehensive. Plus, the explanation language is a bit chatty and conversational, which feels like a human would write, so I gave it a higher credibility score (credibility=0.95)

Input:
<<<
{text}
>>>
Overall verdict (type="{generated verdict type}"): {generated verdict}
\end{tcblisting}

\begin{tcblisting}{
  promptbox,
  label={prompt:win_rate_judge},
  title={Win-rate judge}
}
You are ranking how convincing evidence explanations are.

Task:
Given a source document and {number of candidates} blinded evidence explanations from different annotators, rank all explanations from best to worst according to how convincing the cited evidence and explanation are.

Criteria:
1) Concrete local evidence: specific ideas, elements, or details that are specific to the document.
2) Falsifiability: the explanation should make a specific claim a skeptical reader can verify in the document. It's not enough to say what the evidence is, the explanation should explain why that evidence supports the claim.
3) Internal coherence: the evidence should support the explanation's own AI/human authorship claim without contradicting itself.
4) Human plausibility: a concise forensic observation can beat a polished checklist if it identifies a decisive cue.
5) Real world grounding: the evidence and explanation should fit the document's genre, situation, and general world knowledge.

Important:
- You do not know the true label. Do not use or assume any ground-truth AI/human answer.
- Only rate based on content, not writing style.
- Rank explanation convincingness only from the source document and the candidate explanation.
- You must rank all {number of candidates} items uniquely: rank 1 is most convincing, rank {number of candidates} is least convincing.
- Output JSON only. Do not write any rationale, summary, or commentary outside the ranking array.

Return exactly:
{{
  "ranking": [
    {{"item_id": "A1", "rank": 1, "quality_score": 0.93}},
    ...
  ]
}}

Constraints:
- Include exactly {number of candidates} entries in ranking.
- item_id must match one from the candidate list.
- rank must be integers 1..{number of candidates}, unique.
- quality_score should be float in [0,1], higher is better.

Document:
<document>
{document}
</document>

Candidates (blinded):
{candidates}
\end{tcblisting}

\begin{tcblisting}{
  promptbox,
  label={prompt:win_rate_style_rewrite},
  title={Win-rate style rewrite}
}
Rewrite the comment below so it matches the writing style of the example comment.

Rules:
- Change wording, tone, and sentence flow only. Rewrite the style, but keep the same content.
- Do NOT add, remove, or alter factual claims, cited evidence, or the AI/human conclusion.
- Keep the same evidence points and the same guess direction.
- Do not mention rewriting, style matching, or the example.
- Put the rewritten comment only inside <<< and >>>. Nothing before <<< or after >>>.

Style example (match this voice and cadence, not the facts):
<<<
I think this is human, and I would even guess it comes from a real encyclopedia entry. The facts are very specific: a full birth date, a city, a job title, a war deployment, a time window, and a sports role. The sentence has a real compressed biographical rhythm, with several facts packed into one line. The bracketed citation marks are a strong clue that this was copied from a sourced page, not invented as a smooth paragraph.
>>>

Human comment to rewrite:
<<<
{{human comment}}
>>>
\end{tcblisting}

\FloatBarrier
\clearpage
\section{SFT details}
\label{sec:appendix:sft_details}
We pre-train the policy with supervised fine-tuning (SFT) before GRPO to initialize the
annotation format and label-hint following.
Table~\ref{tab:sft-hyperparams} lists all hyperparameters.

\paragraph{Compute.}
All training was performed on the Tinker platform (Thinking Machines).
SFT ran for 2 epochs ($\approx$\num{1440} examples) and completed in approximately \textbf{2 hours}.
GRPO ran for \textbf{310 steps} with early stopping and completed in approximately \textbf{12 hours}.
Inference for SFT data generation (GPT-5.5 annotations) and win-rate evaluation was performed via the respective model providers' APIs.

\begin{table*}[]
\centering
\caption{SFT pre-training hyperparameters.}
\label{tab:sft-hyperparams}
\begin{tabular}{ll}
\toprule
\textbf{Hyperparameter} & \textbf{Value} \\
\midrule
\multicolumn{2}{l}{\textit{Model}} \\
Base model           & GPT-OSS 120B \\
LoRA rank            & 32 \\
\midrule
\multicolumn{2}{l}{\textit{Optimization}} \\
Optimizer            & Adam \\
Learning rate        & $5 \times 10^{-5}$ \\
Batch size           & 8 \\
Epochs               & 1 \\
Random seed          & 2262 \\
\midrule
\multicolumn{2}{l}{\textit{Data}} \\
Total training examples & $1{,}440$ \\
Loss masking         & Completion-only (prompt tokens have weight 0) \\
\midrule
\multicolumn{2}{l}{\textit{Label-hint injection}} \\
Enabled              & Yes (all examples) \\
Hint mix ratio       & 1.0 \\
Contrastive hint CE  & Yes (correct $+$ flipped-label pair per example) \\
Hint-CE learning rate scale & $2.0\times$ \\
Hint-CE target follow rate (EMA) & 0.99 \\
Hint-CE EMA $\alpha$ & 0.1 \\
\bottomrule
\end{tabular}
\end{table*}

\paragraph{Loss masking.}
Only completion tokens receive a cross-entropy loss weight of 1; all prompt tokens
(instruction, document, and analysis-channel stub) are masked with weight 0.
This ensures the model is supervised solely on the annotation output distribution.

\paragraph{Paced annotation dropout.}
To prevent the model from memorising densely-annotated examples (since some in our dataset have too many annotations clustered together), we stochastically
drop nested span annotations before each training step.
The target density is one annotation per 20 document words; spans with higher
credibility scores (score $= 1.0$) are $3\times$ less likely to be dropped than
low-credibility spans.

\paragraph{Label-hint contrastive CE.}
Each expert-annotated example is paired with a label hint injected into the
analysis channel.
A contrastive cross-entropy auxiliary step is run on each hint: a ``correct hint''
forward pass (hint matches ground-truth label) and a ``flipped hint'' forward pass
(hint is wrong) are both performed, and the model is trained to assign higher
probability to the correct outer annotation type under the correct hint.
This teaches the model to follow label-conditioning signals used during GRPO sampling.

\FloatBarrier
\clearpage
\section{Selection of the annotation format}
\label{sec:appendix:annotation_format}
While we first experimented with an XML-based format (\texttt{<span type="AI|human" why="..." score="0.0">TEXT</span>}), we felt it was conceptually off: we work with causal models, and XML requires that attributes go in the opening tag, which means the model would have to decide how to annotate a span before writing it, rather than writing the text and then annotating it. This is not necessarily problematic since the full text is visible in the input, but we found it cleaner to have a format where the attributes come after the text.

We also considered moving attributes to the closing \texttt{</span>} tag, but we expected that would fight the model's strong conditioning on generating valid XML and hurt performance. A Markdown-inspired format could be an interesting alternative, but the syntax used easily conflicts with natural text, and our tests showed that the model easily collapsed with this paradigm. We finally found the best syntax to be XML-based (\texttt{<span>TEXT<annotation type="AI/human" why="..." score="0.0" /></span>}).

Additionally, we added custom special tokens on the annotation's fixed positions, to decrease the likelihood of format collapse and token usage:
\begin{itemize}
    \item \texttt{<text>} at the start of the text section (emitted only once)
    \item \texttt{<span>} to open each ``tell'' annotation
    \item \texttt{<annotation type="} to be written before closing the annotation
    \item \texttt{" why="} to be written before the explanation
    \item \texttt{" score="} to be written before the credibility score
    \item \texttt{" /></span>} to be written at the end of the annotation
\end{itemize}

It can be observed that these special tokens are not the original XML tokens the model would have observed, so effectively the model is learning a syntax akin to \texttt{[SPECIAL TOKEN]abc[SPECIAL TOKEN]def[SPECIAL TOKEN]AI|human[SPECIAL TOKEN]explanation[SPECIAL TOKEN]score[SPECIAL TOKEN]\dots}.

\FloatBarrier
\clearpage
\section{Detector benchmark details}
\label{sec:appendix:benchmarking_details}
This appendix contains the full detector benchmark results from Section~\ref{sec:results:detector_benchmark}.

\paragraph{Overall ranking (Table~\ref{tab:ranking}).}
The $\dagger$ markers indicate that the gap between \ourmodel~and the next two detectors (MAGE, Pangram EditLens) is not statistically significant, nor are several gaps within the mid-tier cluster at ranks 7--11 (DetectLLM-NPR through LogRank).
Binoculars (\num{0.616}) and DNA-GPT (\num{0.581}) perform substantially below what their original papers report, consistent with the replication gap we discuss in Section~\ref{sec:introduction}.

\begin{table*}[t]
\centering
\caption{Detector ranking on the TELL benchmark test set ($n=\num{5000}$). Failed rows imputed to score $0$. AUROC 95\% CIs from bootstrap resampling ($B=\num{10000}$). Mean Kendall $\tau=\num{0.9753}$. $\dagger$~gap not significant vs.\ adjacent rank below (DeLong, BH FDR $q=0.05$).}
\label{tab:ranking}
\begin{tabular}{r l S[table-format=1.4,detect-weight,mode=text] c S[table-format=1.4,detect-weight,mode=text] r}
\toprule
{Rank} & {Detector} & {AUROC} & {95\% CI} & {TPR@1\%FPR} & {$P(\text{rank holds})$} \\
\midrule
  1 & \bfseries TELL (ours) & \bfseries 0.9270 & [\num{0.9192}, \num{0.9348}] & \bfseries 0.6380 & \num{0.993} \\
  2 & MAGE & 0.9132 & [\num{0.9042}, \num{0.9219}] & 0.0424 & \num{0.647}$\dagger$ \\
  3 & Pangram EditLens & 0.9111 & [\num{0.9028}, \num{0.9191}] & 0.5828 & \num{1.000} \\
  4 & Fast-DetectGPT & 0.8609 & [\num{0.8497}, \num{0.8716}] & 0.5896 & \num{1.000} \\
  5 & ArguGPT & 0.8281 & [\num{0.8164}, \num{0.8397}] & 0.4328 & \num{1.000} \\
  6 & T5Sentinel & 0.8020 & [\num{0.7898}, \num{0.8141}] & 0.1748 & \num{0.985} \\
  7 & DetectLLM-NPR & 0.7824 & [\num{0.7692}, \num{0.7953}] & 0.3196 & \num{0.749}$\dagger$ \\
  8 & OpenAI RoBERTa & 0.7770 & [\num{0.7639}, \num{0.7894}] & 0.3308 & \num{0.622}$\dagger$ \\
  9 & AIGC MPU & 0.7741 & [\num{0.7610}, \num{0.7868}] & 0.1160 & \num{0.899}$\dagger$ \\
  10 & DetectLLM-LRR & 0.7627 & [\num{0.7494}, \num{0.7759}] & 0.2716 & \num{0.957}$\dagger$ \\
  11 & LogRank GPT-2-medium & 0.7573 & [\num{0.7440}, \num{0.7707}] & 0.2320 & \num{0.913}$\dagger$ \\
  12 & RADAR & 0.7441 & [\num{0.7301}, \num{0.7583}] & 0.0128 & \num{1.000} \\
  13 & ChatGPT-D & 0.6972 & [\num{0.6824}, \num{0.7112}] & 0.1660 & \num{1.000} \\
  14 & Binoculars & 0.6162 & [\num{0.6006}, \num{0.6320}] & 0.0140 & \num{0.999} \\
  15 & DNA-GPT & 0.5809 & [\num{0.5662}, \num{0.5954}] & 0.0000 & \num{1.000} \\
  16 & PHD RoBERTa & 0.5206 & [\num{0.5045}, \num{0.5371}] & 0.0460 & \multicolumn{1}{c}{---} \\
\bottomrule
\end{tabular}
\end{table*}

\paragraph{Pairwise significance (Table~\ref{tab:pairwise}).}
\ourmodel's advantage over every detector from rank 4 downward is statistically significant (BH-corrected DeLong test, FDR $q = 0.05$); the only non-bold positive entries in \ourmodel's row are MAGE and Pangram EditLens.
Within ranks 7--11, many pairwise gaps are also non-significant, so the ordering among those detectors should not be over-interpreted.

\begin{table*}[t]
\centering
\caption{Pairwise $\Delta$AUROC (row $-$ col). \textbf{Bold} = BH-significant (DeLong test, FDR $q=0.05$, \num{120} comparisons). Detectors ordered by rank (left/top = best).}
\label{tab:pairwise}
\resizebox{\linewidth}{!}{%
\begin{tabular}{l S[table-format=+1.3,detect-weight,mode=text] S[table-format=+1.3,detect-weight,mode=text] S[table-format=+1.3,detect-weight,mode=text] S[table-format=+1.3,detect-weight,mode=text] S[table-format=+1.3,detect-weight,mode=text] S[table-format=+1.3,detect-weight,mode=text] S[table-format=+1.3,detect-weight,mode=text] S[table-format=+1.3,detect-weight,mode=text] S[table-format=+1.3,detect-weight,mode=text] S[table-format=+1.3,detect-weight,mode=text] S[table-format=+1.3,detect-weight,mode=text] S[table-format=+1.3,detect-weight,mode=text] S[table-format=+1.3,detect-weight,mode=text] S[table-format=+1.3,detect-weight,mode=text] S[table-format=+1.3,detect-weight,mode=text] S[table-format=+1.3,detect-weight,mode=text]}
\toprule
{} & {\textbf{TELL}} & {MAGE} & {Pangram} & {F-DGT} & {ArguGPT} & {T5Sent.} & {DL-NPR} & {OAI-RB} & {AIGC MPU} & {DL-LRR} & {LogRank} & {RADAR} & {CGPT-D} & {Binoc.} & {DNA-GPT} & {PHD} \\
\midrule
  {\textbf{TELL}} & {---} & \bfseries +0.014 & \bfseries +0.016 & \bfseries +0.066 & \bfseries +0.099 & \bfseries +0.125 & \bfseries +0.145 & \bfseries +0.150 & \bfseries +0.153 & \bfseries +0.164 & \bfseries +0.170 & \bfseries +0.183 & \bfseries +0.230 & \bfseries +0.311 & \bfseries +0.346 & \bfseries +0.406 \\
  {MAGE} & \bfseries -0.014 & {---} & +0.002 & \bfseries +0.052 & \bfseries +0.085 & \bfseries +0.111 & \bfseries +0.131 & \bfseries +0.136 & \bfseries +0.139 & \bfseries +0.151 & \bfseries +0.156 & \bfseries +0.169 & \bfseries +0.216 & \bfseries +0.297 & \bfseries +0.332 & \bfseries +0.393 \\
  {Pangram} & \bfseries -0.016 & -0.002 & {---} & \bfseries +0.050 & \bfseries +0.083 & \bfseries +0.109 & \bfseries +0.129 & \bfseries +0.134 & \bfseries +0.137 & \bfseries +0.148 & \bfseries +0.154 & \bfseries +0.167 & \bfseries +0.214 & \bfseries +0.295 & \bfseries +0.330 & \bfseries +0.390 \\
  {F-DGT} & \bfseries -0.066 & \bfseries -0.052 & \bfseries -0.050 & {---} & \bfseries +0.033 & \bfseries +0.059 & \bfseries +0.078 & \bfseries +0.084 & \bfseries +0.087 & \bfseries +0.098 & \bfseries +0.104 & \bfseries +0.117 & \bfseries +0.164 & \bfseries +0.245 & \bfseries +0.280 & \bfseries +0.340 \\
  {ArguGPT} & \bfseries -0.099 & \bfseries -0.085 & \bfseries -0.083 & \bfseries -0.033 & {---} & \bfseries +0.026 & \bfseries +0.046 & \bfseries +0.051 & \bfseries +0.054 & \bfseries +0.065 & \bfseries +0.071 & \bfseries +0.084 & \bfseries +0.131 & \bfseries +0.212 & \bfseries +0.247 & \bfseries +0.307 \\
  {T5Sent.} & \bfseries -0.125 & \bfseries -0.111 & \bfseries -0.109 & \bfseries -0.059 & \bfseries -0.026 & {---} & \bfseries +0.020 & \bfseries +0.025 & \bfseries +0.028 & \bfseries +0.039 & \bfseries +0.045 & \bfseries +0.058 & \bfseries +0.105 & \bfseries +0.186 & \bfseries +0.221 & \bfseries +0.281 \\
  {DL-NPR} & \bfseries -0.145 & \bfseries -0.131 & \bfseries -0.129 & \bfseries -0.078 & \bfseries -0.046 & \bfseries -0.020 & {---} & +0.005 & +0.008 & \bfseries +0.020 & \bfseries +0.025 & \bfseries +0.038 & \bfseries +0.085 & \bfseries +0.166 & \bfseries +0.201 & \bfseries +0.262 \\
  {OAI-RB} & \bfseries -0.150 & \bfseries -0.136 & \bfseries -0.134 & \bfseries -0.084 & \bfseries -0.051 & \bfseries -0.025 & -0.005 & {---} & +0.003 & +0.014 & \bfseries +0.020 & \bfseries +0.033 & \bfseries +0.080 & \bfseries +0.161 & \bfseries +0.196 & \bfseries +0.256 \\
  {AIGC MPU} & \bfseries -0.153 & \bfseries -0.139 & \bfseries -0.137 & \bfseries -0.087 & \bfseries -0.054 & \bfseries -0.028 & -0.008 & -0.003 & {---} & +0.011 & +0.017 & \bfseries +0.030 & \bfseries +0.077 & \bfseries +0.158 & \bfseries +0.193 & \bfseries +0.253 \\
  {DL-LRR} & \bfseries -0.164 & \bfseries -0.151 & \bfseries -0.148 & \bfseries -0.098 & \bfseries -0.065 & \bfseries -0.039 & \bfseries -0.020 & -0.014 & -0.011 & {---} & +0.005 & +0.019 & \bfseries +0.066 & \bfseries +0.146 & \bfseries +0.182 & \bfseries +0.242 \\
  {LogRank} & \bfseries -0.170 & \bfseries -0.156 & \bfseries -0.154 & \bfseries -0.104 & \bfseries -0.071 & \bfseries -0.045 & \bfseries -0.025 & \bfseries -0.020 & -0.017 & -0.005 & {---} & +0.013 & \bfseries +0.060 & \bfseries +0.141 & \bfseries +0.176 & \bfseries +0.237 \\
  {RADAR} & \bfseries -0.183 & \bfseries -0.169 & \bfseries -0.167 & \bfseries -0.117 & \bfseries -0.084 & \bfseries -0.058 & \bfseries -0.038 & \bfseries -0.033 & \bfseries -0.030 & -0.019 & -0.013 & {---} & \bfseries +0.047 & \bfseries +0.128 & \bfseries +0.163 & \bfseries +0.223 \\
  {CGPT-D} & \bfseries -0.230 & \bfseries -0.216 & \bfseries -0.214 & \bfseries -0.164 & \bfseries -0.131 & \bfseries -0.105 & \bfseries -0.085 & \bfseries -0.080 & \bfseries -0.077 & \bfseries -0.066 & \bfseries -0.060 & \bfseries -0.047 & {---} & \bfseries +0.081 & \bfseries +0.116 & \bfseries +0.177 \\
  {Binoc.} & \bfseries -0.311 & \bfseries -0.297 & \bfseries -0.295 & \bfseries -0.245 & \bfseries -0.212 & \bfseries -0.186 & \bfseries -0.166 & \bfseries -0.161 & \bfseries -0.158 & \bfseries -0.146 & \bfseries -0.141 & \bfseries -0.128 & \bfseries -0.081 & {---} & \bfseries +0.035 & \bfseries +0.096 \\
  {DNA-GPT} & \bfseries -0.346 & \bfseries -0.332 & \bfseries -0.330 & \bfseries -0.280 & \bfseries -0.247 & \bfseries -0.221 & \bfseries -0.201 & \bfseries -0.196 & \bfseries -0.193 & \bfseries -0.182 & \bfseries -0.176 & \bfseries -0.163 & \bfseries -0.116 & \bfseries -0.035 & {---} & \bfseries +0.060 \\
  {PHD} & \bfseries -0.406 & \bfseries -0.393 & \bfseries -0.390 & \bfseries -0.340 & \bfseries -0.307 & \bfseries -0.281 & \bfseries -0.262 & \bfseries -0.256 & \bfseries -0.253 & \bfseries -0.242 & \bfseries -0.237 & \bfseries -0.223 & \bfseries -0.177 & \bfseries -0.096 & \bfseries -0.060 & {---} \\
\bottomrule
\end{tabular}}
\end{table*}

\paragraph{Per-domain breakdown (Table~\ref{tab:domain}).}
Aggregate AUROC masks substantial variation across domains.
Some detectors reach near-perfect scores on specific domains --- T5Sentinel achieves \num{1.000} on \texttt{web\_text}, MAGE \num{0.999} on \texttt{commonsense\_completion}, ChatGPT-D \num{1.000} on \texttt{finance} despite ranking 13th overall --- while dropping considerably elsewhere, which suggests domain-specific distributional signals rather than general detection ability.
\ourmodel's weakest domain is \texttt{commonsense\_completion} (\num{0.734}); we attribute this partly to limited coverage of that domain in our training data.
Overall, \ourmodel~is the most consistent detector across domains, with no domain where it substantially underperforms the field.

\begin{table*}[t]
\centering
\caption{Per-domain AUROC on the TELL benchmark test set. Best result per domain in \textbf{bold}. Detectors ordered by overall rank (left = best).}
\label{tab:domain}
\resizebox{\linewidth}{!}{%
\begin{tabular}{l S[table-format=1.3,detect-weight,mode=text] S[table-format=1.3,detect-weight,mode=text] S[table-format=1.3,detect-weight,mode=text] S[table-format=1.3,detect-weight,mode=text] S[table-format=1.3,detect-weight,mode=text] S[table-format=1.3,detect-weight,mode=text] S[table-format=1.3,detect-weight,mode=text] S[table-format=1.3,detect-weight,mode=text] S[table-format=1.3,detect-weight,mode=text] S[table-format=1.3,detect-weight,mode=text] S[table-format=1.3,detect-weight,mode=text] S[table-format=1.3,detect-weight,mode=text] S[table-format=1.3,detect-weight,mode=text] S[table-format=1.3,detect-weight,mode=text] S[table-format=1.3,detect-weight,mode=text] S[table-format=1.3,detect-weight,mode=text]}
\toprule
{Domain} & {\textbf{TELL}} & {MAGE} & {Pangram} & {F-DGT} & {ArguGPT} & {T5Sent.} & {DL-NPR} & {OAI-RB} & {AIGC MPU} & {DL-LRR} & {LogRank} & {RADAR} & {CGPT-D} & {Binoc.} & {DNA-GPT} & {PHD} \\
\midrule
  {academic\_abstract} & \bfseries 0.971 & 0.970 & 0.942 & 0.860 & 0.862 & 0.880 & 0.759 & 0.747 & 0.762 & 0.783 & 0.738 & 0.782 & 0.735 & 0.515 & 0.574 & 0.503 \\
  {commonsense\_completion} & 0.734 & \bfseries 0.999 & 0.500 & 0.689 & 0.506 & 0.367 & 0.581 & 0.522 & 0.418 & 0.646 & 0.654 & 0.336 & 0.487 & 0.320 & 0.600 & 0.703 \\
  {creative\_writing} & \bfseries 0.928 & 0.894 & 0.908 & 0.855 & 0.812 & 0.658 & 0.805 & 0.703 & 0.838 & 0.800 & 0.805 & 0.701 & 0.625 & 0.628 & 0.618 & 0.515 \\
  {educational\_web} & 0.993 & 0.840 & \bfseries 1.000 & 0.965 & 0.926 & 0.877 & 0.863 & 0.759 & 0.947 & 0.760 & 0.862 & 0.956 & 0.631 & 0.809 & 0.608 & 0.409 \\
  {email} & 0.998 & 0.921 & \bfseries 1.000 & 0.919 & 0.989 & 0.625 & 0.956 & 0.547 & 0.991 & 0.931 & 0.943 & 0.740 & 0.884 & 0.733 & 0.492 & 0.518 \\
  {encyclopedic\_reference} & 0.911 & 0.902 & 0.899 & 0.888 & 0.782 & 0.888 & 0.785 & 0.877 & 0.642 & 0.794 & 0.764 & \bfseries 0.913 & 0.724 & 0.632 & 0.576 & 0.548 \\
  {finance} & 0.964 & 0.987 & 0.989 & 0.986 & 0.997 & 0.605 & 0.987 & 0.986 & 0.919 & 0.976 & 0.981 & 0.647 & \bfseries 1.000 & 0.779 & 0.798 & 0.842 \\
  {forum\_qa} & 0.921 & \bfseries 0.984 & 0.924 & 0.887 & 0.864 & 0.808 & 0.878 & 0.834 & 0.723 & 0.870 & 0.859 & 0.781 & 0.803 & 0.637 & 0.594 & 0.544 \\
  {howto\_instructional} & 0.896 & 0.730 & \bfseries 0.934 & 0.879 & 0.720 & 0.845 & 0.745 & 0.798 & 0.734 & 0.697 & 0.702 & 0.606 & 0.558 & 0.676 & 0.532 & 0.634 \\
  {news} & 0.901 & \bfseries 0.966 & 0.913 & 0.867 & 0.817 & 0.847 & 0.744 & 0.807 & 0.823 & 0.730 & 0.715 & 0.838 & 0.670 & 0.625 & 0.554 & 0.435 \\
  {review\_opinion} & 0.930 & \bfseries 0.974 & 0.905 & 0.846 & 0.858 & 0.822 & 0.816 & 0.782 & 0.768 & 0.789 & 0.784 & 0.748 & 0.729 & 0.649 & 0.592 & 0.427 \\
  {student\_essay} & 0.993 & 0.952 & \bfseries 0.997 & 0.914 & 0.990 & 0.948 & 0.967 & 0.945 & 0.979 & 0.909 & 0.941 & 0.919 & 0.787 & 0.809 & 0.632 & 0.656 \\
  {web\_text} & 0.785 & 0.609 & 0.804 & 0.598 & 0.538 & \bfseries 1.000 & 0.352 & 0.757 & 0.485 & 0.430 & 0.335 & 0.860 & 0.554 & 0.425 & 0.469 & 0.465 \\
\bottomrule
\end{tabular}}
\end{table*}

\paragraph{Reward hacking.} Also, while training, we observed that the model sometimes learns to ``hack'' the reward function by making the annotations seem more credible --- through deception. For instance, the model learned to write ``this is a very common strong AI sign'' as a suffix to all its AI spans, which caused the judge to give higher rewards to all those spans, even though there was essentially no extra information in that phrase. We also saw cases of the model lying and saying e.g. ``I’ve seen the way the author uses punctuation in other texts, and this shows a similar pattern''. We adjusted our judge prompt to base its reward on the quality of the explanation exclusively, ignoring external factors that cannot be verified.

\FloatBarrier
\clearpage
\section{Win-Rate evaluation}
\label{sec:win_rate_evaluation}
Here, we share additional details on how we designed and performed our win-rate evaluations.

We evaluate explanation quality with a blinded listwise judge study on the TELL human-detectors validation set (to reduce computational costs). For each of 200 documents, we sample one model explanation from the trained policy and compare it to five human annotator explanations. We style-normalize the human comments (see below) so judges compare their content rather than surface form. We thus present the judges with six candidates in a random blind order, and each ``member'' of a five-judge panel (\texttt{GPT-5.4-mini}, \texttt{DeepSeek-V4-Flash}, \texttt{NVIDIA-Nemotron-3-Super-120B-A12B}, \texttt{gemma-4-26b} and \texttt{GPT-OSS-120B}) produces a full ranking from most to least convincing.

From each ranking we derive a document-level win rate: the fraction of pairwise model-human comparisons the model wins within that document (ties count as 0.5). Our primary metric is the panel document win rate, the mean of these per-document rates averaged across judges. In inference, we treat the document as the unit of analysis: we test whether the panel mean exceeds 0.5 (no aggregate advantage) with a one-sided sign-flip permutation test and a one-sided Wilcoxon signed-rank test on per-document panel scores, and we report a 95\% confidence interval by bootstrapping the documents. We report the per-judge results as a robustness check (with Holm-adjusted p-values to account for testing multiple judges). We show the complete results in Table~\ref{tab:winrate}.

\paragraph{Self-preference bias.} With the recent body of research showing that LLMs tend to prefer polished, ``AI-sounding'' text in mind \cite{lauritoAIAIBias2025,kooBenchmarkingCognitiveBiases2024,bittonVisITBenchBenchmarkVisionLanguage2023,liuLLMsNarcissisticEvaluators2024}, we thought it potentially biased to put human-style comments and AI-generated text in the same evaluation, since the LLM judge might be more likely to prefer the latter only based on the surface-level style rather than the content or accuracy of the information presented. Therefore, we re-wrote all the human comments with the same model used for \ourmodel, so that their content would remain intact but their style would be more similar to \ourmodel's outputs (see prompt \ref{prompt:win_rate_style_rewrite}). This way, we aimed to minimize the influence of style in the evaluation so that win-rate is focused on the actual quality of the explanations themselves.

\begin{table*}[t]
\centering
\caption{Listwise win rate vs.\ human annotators (TELL human-detectors test, $n=\num{200}$ documents with complete 5-judge panel). 95\% CIs: document-level bootstrap ($B=\num{10000}$). $p_{\mathrm{perm}}$: one-sided sign-flip permutation vs.\ 50\%; $p_{\mathrm{Wilc}}$: one-sided Wilcoxon signed-rank vs.\ 50\%.}
\label{tab:winrate}
\begin{tabular}{l S[table-format=2.1] c c c}
\toprule
{Judge} & {Win rate (\%)} & {95\% CI} & {$p_\mathrm{perm}$} & {$p_\mathrm{Wilc}$} \\
\midrule
\bfseries Panel mean & \bfseries \num{72.3} & [\num{68.3}, \num{76.2}] & $<10^{-4}$ & $<10^{-4}$ \\
\midrule
  GPT-5.4-mini & \num{78.3} & [\num{73.9}, \num{82.4}] & $<10^{-4}$ & $<10^{-4}$ \\
  Gemma 4 26B & \num{67.5} & [\num{62.6}, \num{72.1}] & $<10^{-4}$ & $<10^{-4}$ \\
  DeepSeek V4 Flash & \num{75.3} & [\num{70.8}, \num{79.5}] & $<10^{-4}$ & $<10^{-4}$ \\
  Nemotron Super & \num{66.3} & [\num{61.5}, \num{70.8}] & $<10^{-4}$ & $<10^{-4}$ \\
  GPT-OSS 120B & \num{74.1} & [\num{69.5}, \num{78.4}] & $<10^{-4}$ & $<10^{-4}$ \\
\bottomrule
\end{tabular}
\end{table*}

\subsection{Example of one of the evaluations.}

Here, we show a specific example to illustrate how the win-rate evaluation works in practice.
We keep the original JSON structure because we believe it makes the examples simpler to understand.

The first part of the pipeline rewrites the human annotations into a more standard format to diminish the influence of writing style on the evaluation:

\begin{tcblisting}{
  box,
  title={Evaluation example before rewriting}
}
"annotator_1": {
  "comment": "The quotes in this piece of text come across as very natural. Some fit in well, others need surrounding text to make sense, and the sources are varied. The professors are also not referred to as doctors, even though they're probably qualified enough. There's also just an absence of anything that might point towards it being AI-generated. ",
  "confidence": 4,
  "guess": "Human-Generated"
},
"annotator_2": {
  "comment": "There are none of the usual AI-repeated words. There is a wider range of vocabulary than usual for AI.",
  "confidence": 4,
  "guess": "Human-Generated"
},
"annotator_3": {
  "comment": "Here's why I think it's human-generated: Instead of saying things like 'researchers don't agree on' AI would have said 'researchers disagree on'. Missing punctuation marks. ",
  "confidence": 5,
  "guess": "Human-Generated"
},
"annotator_4": {
  "comment": "While the writing is more simplistic, it's able to convey the topic well. It uses dashes, colons, and commas to intersperse information with quotes, works with simple, easy to understand phrases such as \"a leg up compared to all other species.\" and \"Pagel, on the other hand, is less sure about hand gestures.\" as a way to explain information in a readable format. It even adds unique phrases, such as \"a small repertoire of sounds and signals with various meanings\" to add to the content. While there's not much sentence variety and creative flair, it doesn't appear AI-generated because it keeps its information concise. So, it's human-written. ",
  "confidence": 4,
  "guess": "Human-Generated"
},
"annotator_5": {
  "comment": "Extensive use of personal pronouns. Quotations provide detail. Distinct variation in sentence and paragraph length. Highlighted sentence, for e.g., is 40 words. Rare in machine-generated text territory. Use of idiom in introductory sentence.",
  "confidence": 5,
  "guess": "Human-Generated"
}
\end{tcblisting}

After rewriting the original human annotations into a more standardized format, we run the win-rate evaluation with the following input (note that A3 is generated by \ourmodel):

\begin{tcblisting}{
  box,
  title={Evaluation example}
}
[
  {
    "item_id": "A1",
    "text": "The quotations feel strikingly natural. Some blend seamlessly, while others rely on additional context to click, and the references are all over the map. The instructors aren’t called doctors, even though they’re likely qualified. Moreover, there’s nothing in the passage that hints at AI authorship."
  },
  {
    "item_id": "A2",
    "text": "Here's why I see it as human-made: it avoids the phrasing “researchers don’t agree on,” which an AI would likely condense to “researchers disagree on.” It also drops the expected punctuation."
  },
  {
    "item_id": "A3",
    "text": "I think this is human-written. The piece is very smooth and conversational, but it has a real magazine rhythm. I see a lot of human choices in the phrasing, like the casual opener, the little aside about recording devices, and the idiom “leg up”. The quotes are woven into the story in different ways, sometimes with “says” and sometimes with “adds”. There are a few repeated phrases, like the whole gesture-vocalization sentence, which feel like a human drafting or editing slip. The sentence structure is varied, with long explanatory clauses and short punchy lines, and the writer uses small asides and dashes to control the pace."
  },
  {
    "item_id": "A4",
    "text": "It lacks the typical repetitive phrasing you often see from AI, and it actually employs a broader vocabulary than what AI usually produces."
  },
  {
    "item_id": "A5",
    "text": "The text leans heavily on personal pronouns and peppers in quotations for detail. It shows a clear mix of sentence and paragraph lengths, even featuring a highlighted line that stretches to about forty words—something you don’t often see in AI‑produced prose. Plus, it kicks off with an idiom, adding a human‑like touch."
  },
  {
    "item_id": "A6",
    "text": "The prose is straightforward yet manages to get the point across. It leans on dashes, colons, and commas to weave in quotes, using plain, clear phrasing like “a leg up compared to all other species” and “Pagel, on the other hand, is less sure about hand gestures” to explain the ideas in an accessible way. It even slips in distinctive wording such as “a small repertoire of sounds and signals with various meanings” to enrich the text. Although the sentence structure isn’t especially varied or flamboyant, the concise delivery suggests it isn’t AI‑generated. In short, it reads as human‑written."
  }
]
\end{tcblisting}

After evaluation, these are the results obtained (\ourmodel ranks first, followed by A6, since it is more specific and detailed):

\begin{tcblisting}{
  box,
  title={Reasoning}
}
"ranking": [
  {
    "item_id": "A3",
    "rank": 1,
    "quality_score": 0.86,
    "original_item_id": "M1",
    "true_kind": "model",
    "source_id": "model_rollout_1"
  },
  {
    "item_id": "A6",
    "rank": 2,
    "quality_score": 0.74,
    "original_item_id": "H4",
    "true_kind": "human",
    "source_id": "annotator_4"
  },
  {
    "item_id": "A2",
    "rank": 3,
    "quality_score": 0.41,
    "original_item_id": "H3",
    "true_kind": "human",
    "source_id": "annotator_3"
  },
  {
    "item_id": "A5",
    "rank": 4,
    "quality_score": 0.28,
    "original_item_id": "H5",
    "true_kind": "human",
    "source_id": "annotator_5"
  },
  {
    "item_id": "A4",
    "rank": 5,
    "quality_score": 0.16,
    "original_item_id": "H2",
    "true_kind": "human",
    "source_id": "annotator_2"
  },
  {
    "item_id": "A1",
    "rank": 6,
    "quality_score": 0.08,
    "original_item_id": "H1",
    "true_kind": "human",
    "source_id": "annotator_1"
  }
],
"short_rationale": "A3 is strongest because it cites several concrete, document-specific features: the casual opener, the repeated gesture/vocalization sentence, the varied quote attributions, and the pacing created by dashes and short/long sentences. A6 is also grounded in actual wording from the passage, but it mostly praises clarity and style rather than identifying a decisive inconsistency or error. A2 is weaker because it hinges on a speculative AI-vs-human paraphrase and a vague punctuation claim. A5 and A4 rely on generic detector folklore about pronouns, sentence length, vocabulary, and repetition, with little falsifiable evidence. A1 is least convincing because it is largely disconnected from the document and makes unsupported claims about instructors/doctors that do not appear in the passage."
\end{tcblisting}

\end{document}